\documentclass{article}

\usepackage{microtype}
\usepackage{graphicx}
\usepackage{subfigure}
\usepackage{booktabs} %
\usepackage{tikz}
\usepackage{hyperref}

\usepackage[accepted]{icml2024}
\usepackage{flushend}

\usepackage{xcolor}         %
\usepackage{amsmath}
\usepackage{amssymb}
\usepackage{mathtools}
\usepackage{amsthm}
\usepackage{amsfonts}
\usepackage{bm}
\usepackage{enumitem}
\usepackage{graphicx}
\usepackage{float}
\usepackage{adjustbox}
\usepackage{wrapfig}
\usepackage[symbol]{footmisc}
\usepackage{subcaption}
\usepackage{cases}
\usepackage{dsfont}
\usepackage{subcaption}
\usepackage{array}
\usepackage{multirow}
\usepackage{multicol}
\usepackage{stfloats} 

\definecolor{anthracite}{RGB}{13, 13, 21}
\definecolor{indigo}{RGB}{63, 81, 181}
\definecolor{red}{RGB}{210, 40, 95} 
\definecolor{pink}{RGB}{236, 64, 122}
\definecolor{green}{RGB}{46, 182, 125}
\definecolor{blue}{RGB}{66, 133, 244}
\definecolor{yellow}{RGB}{236, 178, 46}
\definecolor{anthracite}{RGB}{13, 13, 21}
\definecolor{gold}{RGB}{182, 131, 76}
\definecolor{lightgrey}{RGB}{128, 128, 128}

\newcommand{\best}[1]{\textbf{\textcolor{anthracite}{#1}}}
\newcommand{\forgrad}{\textbf{\textcolor{anthracite}{FORGrad}}}

\newcommand{\pred}{\bm{f}}

\newcommand{\F}{\mathcal{F}}

\def\sX{{\mathcal{X}}}
\def\sY{{\mathcal{Y}}}
\def\vx{{\bm{x}}}

\newcommand{\x}{\bm{x}}
\newcommand{\am}{\bm{\varphi}}

\newcommand{\ep}{\bm{\varepsilon}}

\DeclareMathOperator*{\argmax}{arg\,max}

\usepackage[capitalize,noabbrev]{cleveref}

\theoremstyle{plain}

\theoremstyle{definition}

\theoremstyle{remark}

\usepackage[textsize=tiny]{todonotes}

\icmltitlerunning{Saliency strikes back: How filtering out high frequencies improves white-box explanations}

\begin{document}

\twocolumn[
\icmltitle{Saliency Strikes Back:
How Filtering out High Frequencies Improves White-Box Explanations}

\icmlsetsymbol{equal}{*}

\begin{icmlauthorlist}
\icmlauthor{Sabine Muzellec}{yyy,comp}
\icmlauthor{Thomas Fel}{yyy,sncf}
\icmlauthor{Victor Boutin}{yyy,comp}
\icmlauthor{Léo Andéol}{sch,sncf}
\icmlauthor{Rufin VanRullen}{comp}
\icmlauthor{Thomas Serre}{yyy}
\end{icmlauthorlist}

\icmlaffiliation{yyy}{Carney Institute for Brain Science, Brown University, USA}
\icmlaffiliation{comp}{CerCo, CNRS, France}
\icmlaffiliation{sch}{Institute of Mathematics of Toulouse, University Paul Sabatier, France}
\icmlaffiliation{sncf}{SNCF, France}

\icmlcorrespondingauthor{Sabine Muzellec}{sabine\_muzellec@brown.edu}

\icmlkeywords{Explainability, Attribution methods, Fourier, Deep Learning, Computer Vision}
\vskip 0.3in
]

\printAffiliationsAndNotice{}  %

\begin{abstract}

Attribution methods correspond to a class of explainability methods (XAI) that aim to assess how individual inputs contribute to a model's decision-making process. We have identified a significant limitation in one type of attribution methods, known as ``white-box" methods. Although highly efficient, as we will show, these methods rely on a gradient signal that is often contaminated by high-frequency artifacts. To overcome this limitation, we introduce a new approach called ``\forgrad.'' This simple method effectively filters out these high-frequency artifacts using optimal cut-off frequencies tailored to the unique characteristics of each model architecture. Our findings show that \forgrad~\textit{consistently enhances} the performance of existing white-box methods, enabling them to compete effectively with more accurate yet computationally more demanding ``black-box" methods. We anticipate that, because of its effectiveness, the proposed method will foster the broader adoption of straightforward and efficient white-box methods for explainability, providing a better balance between faithfulness and computational efficiency.

\end{abstract}

\section{Introduction}
Understanding how modern Artificial Neural Networks (ANNs) make decisions remains a major challenge. Given the ever-increasing range of machine learning applications, the need for robust and reliable explainability methods is pressing~\cite{doshi2017towards,jacovi2021formalizing}. To meet this demand, eXplainable Artificial Intelligence (XAI) focuses on developing new tools to help users better understand how ANNs arrive at their decisions~\cite{jacovi2021formalizing,doshivelez2017rigorous,rudin2019stop}. Among the wide range of XAI approaches, attribution methods have become the go-to approaches. They aim to identify the features that drive a network's decisions~\cite{simonyan2013deep} by assigning them an importance score according to their contribution to the overall prediction.
\enlargethispage{3mm}
Attribution methods fall broadly into two main classes: ``white-box'' and ``black-box'' methods. White-box methods require full access to a network's information (weights, architecture, etc.) to understand which regions in the input are the most important for the model's decision. Typically, such methods rely on the gradient of the model's prediction w.r.t. the input to produce attribution maps. For instance, ``Saliency'', one of the earliest white-box methods, traces a model's decision back to its input image using such the gradient information~\cite{simonyan2013deep}. Although simple and computationally efficient, Saliency is also known to produce noisy and hard-to-interpret attribution maps~\cite{nguyen2021effectiveness,kim2021hive,fel2021cannot,serrurier2022adversarial,hase2020evaluating}. Several extensions have been proposed to try to overcome these limitations, e.g., by varying slightly the input and accumulating the resulting gradient to obtain a smoother map~\cite{simonyan2013deep,zeiler2014visualizing,springenberg2014striving,smilkov2017smoothgrad,sundararajan2017axiomatic,zhou2016learning,Selvaraju_2019,Fong_2017}.

Conversely, black-box methods focus on a model's predictions without explicitly requiring any gradient computation. Such methods aim to quantify how changes in the inputs affect a model's output by altering the input and evaluating the newly obtained prediction~\cite{zeiler2014visualizing,petsiuk2018rise,fel2021look,ribeiro2016i,novello2022making}. These black-box methods are computationally much more demanding as they require hundreds of network queries (or forward passes), but they outperform white-box methods on current XAI benchmarks. We refer the reader to Section~\ref{sec:related_work} for more details on the different types of attribution methods. Despite extensive research~\cite{sixt2020explanations,adebayo2018sanity,slack2021reliable,hanexplanation,agarwal2022rethinking,agarwal2021towards,pukdee2023learning}, a question remains:  Why do black-box methods outperform white-box methods?

\begin{figure*}[h]
\center
\begin{tikzpicture}
\draw [anchor=north west] (0\linewidth, 1\linewidth) node {\includegraphics[width=0.45\linewidth]{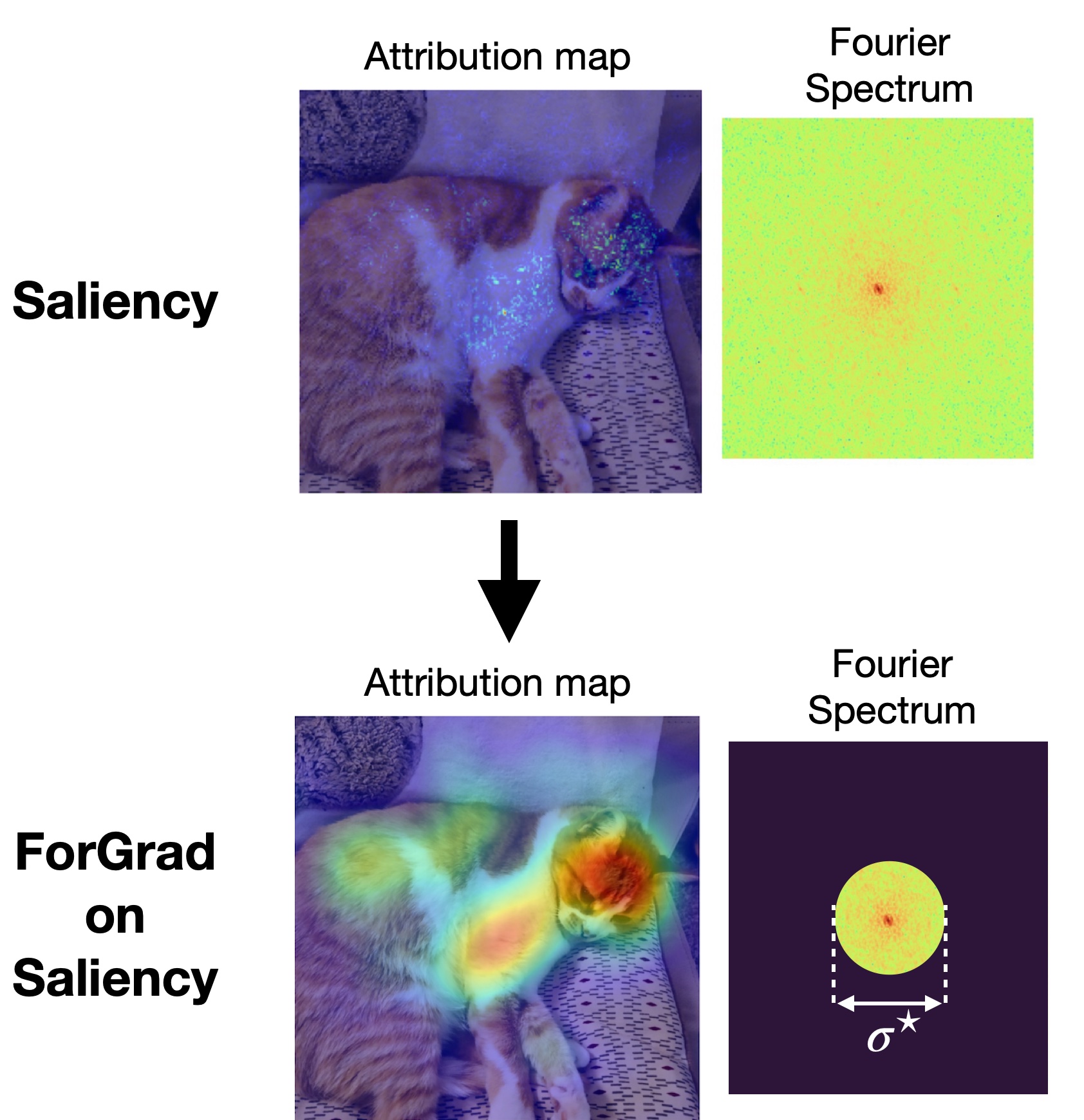}};
\draw [anchor=north west] (0.5\linewidth, 1\linewidth) node {\includegraphics[width=0.5\linewidth]{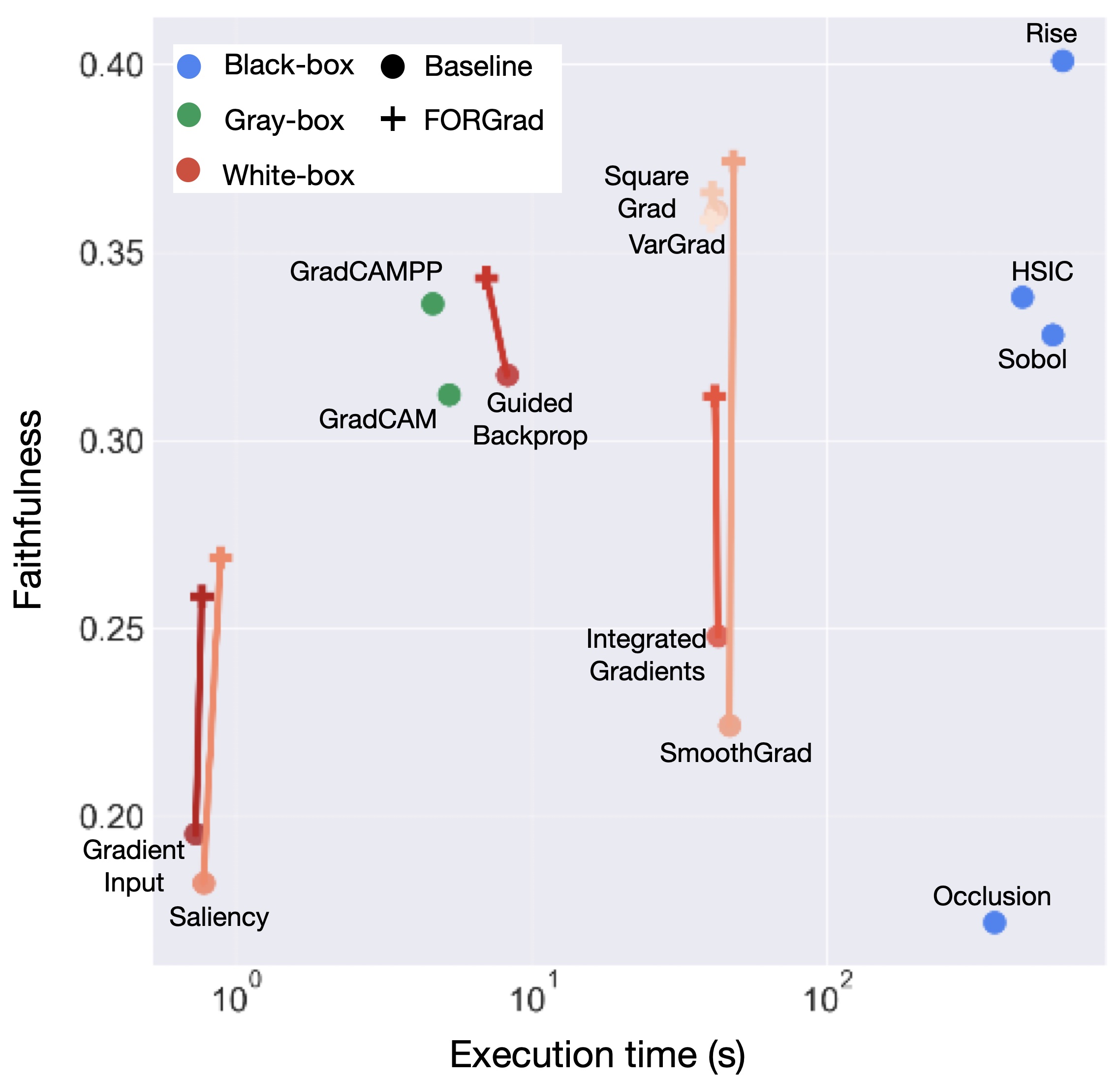}};
\begin{scope}
\draw [anchor=north west,fill=white, align=left] (0.0\linewidth, 1\linewidth) node {\bf A };
\draw [anchor=north west,fill=white, align=left] (0.5\linewidth, 1\linewidth) node {\bf B};
\end{scope}
\end{tikzpicture}
\caption{\textbf{The \forgrad~method.} {\bf A:} \forgrad~estimates an optimal cut-off frequency ($\sigma^{\star}$) for individual combinations of attribution method and network architecture. It is then used to low-pass filter the gradient signal. \forgrad~leads to quantitatively better attribution maps. {\bf B:} \forgrad~is applicable to all white-box methods (reddish data points), and it consistently improves their faithfulness (see reddish crosses). The x-axis represents the execution time in seconds for each method computed on $100$ images. We refer the reader to section~\ref{sec:metrics} for more details on the faithfulness metric.}
\label{fig:forgrad_methods}
\end{figure*}

Consistent with findings from multiple studies~\cite{, springenberg2014striving, selvaraju2017grad,ghorbani2017interpretation,ancona2017better, balduzzi2017shattered,
adebayo2018sanity, kim2019saliency, tomsett2019sanity, rao2022towards}, we have found empirically that white-box methods typically produce attribution maps with increased power in the high frequencies relative to their black-box counterparts. Our analysis further reveals that these high frequencies are inherited from the gradient itself. 
We have also found that these high frequencies often arise due to some downsampling operations within the network, such as max-pooling or striding. This observation suggests that white-box methods could be ``repaired'' by filtering out the gradient signal with carefully chosen network-specific cut-off frequencies. %

Here, we introduce~\forgrad~ FOurier Reparation of the Gradients, a simple method applicable to all white-box methods with negligible computational overhead. \forgrad~filters high-frequencies in the gradient using a cut-off frequency adapted for the network architecture and the attribution method used. This frequency is precomputed on training images to maximize the faithfulness of individual attribution methods. Experimentally, we demonstrate that \forgrad~consistently improves all existing white-box attribution methods. The method and its effects are summarized in Fig.~\ref{fig:forgrad_methods}. 

Overall, \forgrad~allows white-box methods to compete with the computationally demanding black-box methods on XAI benchmarks. Our contributions are summarized as follows:\vspace{-3mm}
\begin{itemize}
    \item We found that a major source of high-frequency artifacts in attribution maps computed with white-box methods is inherited from the model gradient. These artifacts are responsible for the lower explainability scores of these methods.
    \item These gradient artifacts are a consequence of the max-pooling and striding operations used in convolutional neural networks (CNNs).
    \item We introduce \forgrad~to repair white-box methods. \forgrad~filters-out frequencies above a certain cut-off value, which is optimized for a given network architecture. \forgrad~systematically improves the explainability score of white-box methods to levels where they now compete with the computationally intensive black-box methods.
\end{itemize}

\newpage

\section{Related Work}\label{sec:related_work}
\textbf{Black-box attribution methods: }
In black-box attribution methods, the analytical form and potential internal states of the model are unknown. Therefore, those methods exclusively rely on the network's predictions. The first method, Occlusion~\cite{zeiler2014visualizing}, masks individual image regions, one at a time, using an occluding mask set to a baseline value. The corresponding prediction scores are assigned to all pixels within the occluded region, providing an easily interpretable explanation. However, occlusion falls short in capturing the (higher-order) interactions among various image regions -- also called coalitions~\cite{idrissi2023coalitional}. 
Sobol~\cite{fel2021look}, along with related methods such as LIME~\cite{ribeiro2016i} and RISE~\cite{petsiuk2018rise}, address this problem by randomly perturbing multiple regions of the input image simultaneously. Surprisingly, methods like RISE~\cite{petsiuk2018rise} Sobol~\cite{fel2021look} and HSIC~\cite{novello2022making} have outperformed their white-box counterparts which have full access to network inner information (see Fig.~\ref{fig:forgrad_methods}B and Fig.~\ref{fig:barchart}B). However, they come with a trade-off in terms of speed: black-box methods typically need thousands of network queries to generate a comprehensive explanation (see Fig.~\ref{fig:barchart}B).

\textbf{White-box attribution methods: }In contrast to black-box methods, white-box methods have full access to the network information (e.g., the weights, the architecture) and rely on finer information contained in the gradient signal. It was the original idea of the Saliency method~\cite{simonyan2013deep}, the first to back-propagate the gradient from the output to the network's input, indicating which pixels affect the decision score the most. Although computationally very efficient, Saliency is limited because it focuses on the influence of individual pixels in an infinitesimal neighborhood in the input image. For instance, it has been shown that gradients often vanish when the prediction score to be explained is near the maximum value~\cite{sundararajan2017axiomatic,miglani2020investigating}. Integrated Gradient~\cite{sundararajan2017axiomatic}, SmoothGrad~\cite{smilkov2017smoothgrad}, and SquareGrad \cite{hooker2018benchmark,seo2018noise} partially address this issue by accumulating gradients. However, such white-box methods still lag behind black-box attribution methods.

\textbf{Gray-box attribution methods: } The distinction between white-box and black-box methods is nuanced, particularly as some approaches leverage only partial knowledge of the network. Grad-CAM~\cite{Selvaraju_2019} and Grad-CAM++~\cite{chattopadhay2018grad} fall within this category. These methods use the gradient information only of the last layers of the network, as opposed to the full back-propagation done by white-box methods.
Specifically, Grad-CAM and Grad-CAM++ generate visual explanations for class labels using a weighted combination of the positive partial derivatives of the last convolutional layer feature maps with respect to a specific class score. This process generates small sample-specific maps that are then upsampled to fit the input size. Although these methods perform relatively well and are computationally efficient, the corresponding attribution maps often lack fine details~\cite{nam2021interpreting}. This limitation primarily stems from the low-spatial resolution gradient maps they rely on to produce attribution maps. 

\section{Notations, Metrics, and Networks}\label{sec:metrics}
\textbf{Notations: } 
We consider a general supervised learning setting, where a classifier $\pred : \sX \to \sY$ maps images from an input space $\sX \subseteq \mathbb{R}^{W \times H}$ to an output space $\sY \subseteq \mathbb{R}$. %
We denote $\F$ %
the Discrete Fourier Transform (DFT) on $\mathbb{R}^{W\times H}$. %
Importantly, all Fourier spectra shown in this article are symmetrized, i.e., we always shift the low-frequency components to the center of the spectrum.  
We recall that an attribution method is a function $\am : \sX \to \mathbb{R}^{W \times H}$ that
maps an input of interest to its corresponding importance scores $\am(\vx)$. Finally, we denote by $\am_\sigma(\vx)$ the attribution method in which high frequencies are filtered out using a cut-off value of $\sigma$.

\begin{figure}[h!]
\center
\includegraphics[width=0.45\textwidth]{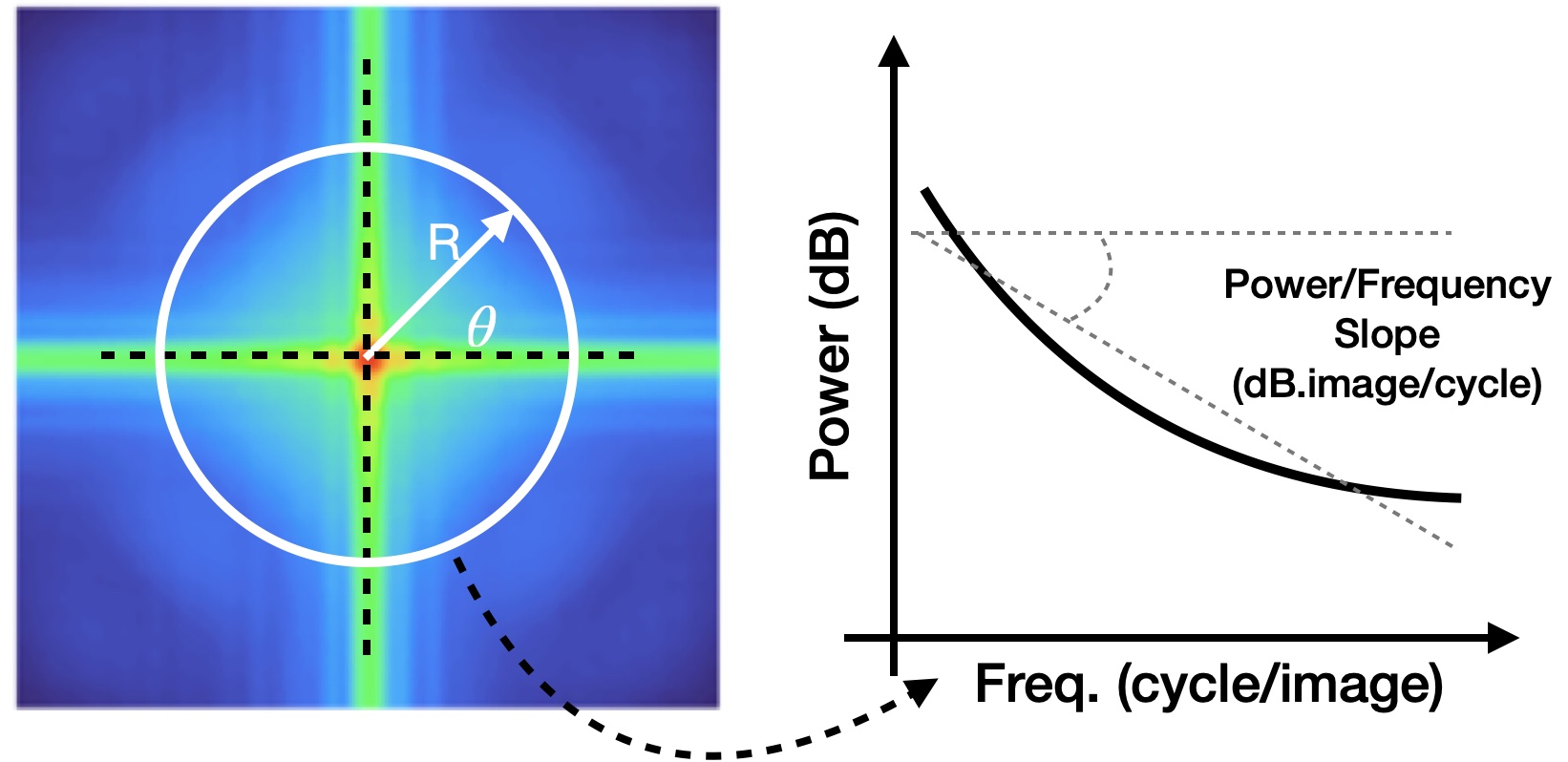}
\caption{\textbf{Fourier signature and power/frequency slope.} The Fourier signature is computed using a circular average of the Fourier amplitudes (averaged over all $\theta$ values) at each frequency (i.e., each $R$). The power/frequency slope summarizes in one scalar the Fourier signature.}
\label{fig:methods}
\end{figure}

\textbf{Fourier signature and power/frequency slope: } In this article, we analyze the Fourier signature of attribution maps produced by white-box and black-box methods. To do so, we compute the amplitude of the Fourier spectrum, $|(\F \circ \am)(\vx)|$ for attribution maps obtained with different methods. We average the spectrum over $1,000$ images. In Fig.~\ref{fig:methods}, we illustrate how to extract the average Fourier signature from the Fourier spectrum computed over attribution (or gradient) maps. This signature is computed using the circular average of the Fourier amplitudes (averaged over all $\theta$ values) at each frequency (i.e., each $R$). We then derive a single scalar, called the power/frequency slope, as the slope of the best linear fit of the Fourier signature. Intuitively, a steeply negative slope indicates a low power in the high-frequencies. Inversely, a slope close to zero suggests that the underlying attribution maps contain a lot of high-frequency signals.

\textbf{XAI metrics: } To evaluate XAI performance, we use the faithfulness metric, denoted $F(\x, \am)$, which is a measure that includes the Insertion and Deletion metrics~\cite{petsiuk2018rise}. Those metrics are widely used in XAI to evaluate the quality of attribution methods~\cite{Fong_2017}. The Deletion metric, $D(\x, \am)$, measures the evolution of the prediction probability when one incrementally removes pixels by replacing them with a baseline value according to their attribution score. Insertion, $I(\x, \am)$, inserts pixels gradually into a baseline input. The faithfulness is obtained using $F(\x, \am) = I(\x, \am) - D(\x, \am)$. Note that the faithfulness metric has shown to be effective in evaluating attribution methods~\cite{samek2016evaluating,li2022fd}. Additionally, we also evaluate attribution methods using the $\mu$Fidelity, denoted $\mu$F~\cite{bhatt2020evaluating}. The fidelity metric serves to verify the correlation between the attribution score and a random subset of pixels. To achieve this, a set of pixels is randomly chosen and set to a baseline state, after which a prediction score is obtained. The fidelity metric evaluates the correlation between the decrease in the score and the significance of the explanation for each random subset created. 
Finally, the sensitivity metric, $S(\x, \am)$~\citep{bhatt2020evaluating}, measures a method's stability by calculating the distance between two explanations. These explanations are generated by the same method, with the first using the original input and the second using the same input perturbed with random noise. A good method, according to this metric, should produce similar explanations for the two inputs, therefore obtaining a low sensitivity score.

\textbf{Networks: }We analyze attribution methods on three different representative ImageNet-trained classifiers: ResNet50V2~\cite{he2016deep}, ConvNeXt~\cite{Liu_2022_CVPR}, and ViT~\cite{dosovitskiy2020image}. For concision, we include in the main text only the results obtained with the ResNet50V2 network and we refer the reader to supplementary information for the $2$ other networks. All the results obtained on ConvNeXt and ViT are consistent with those of ResNet50V2.

\section{White-box Methods are Contaminated by High-Frequencies from the Gradient}

\begin{figure*}[h!]
\center
\begin{tikzpicture}
\draw [anchor=north west] (0\linewidth, 1\linewidth) node {\includegraphics[width=0.42\linewidth]{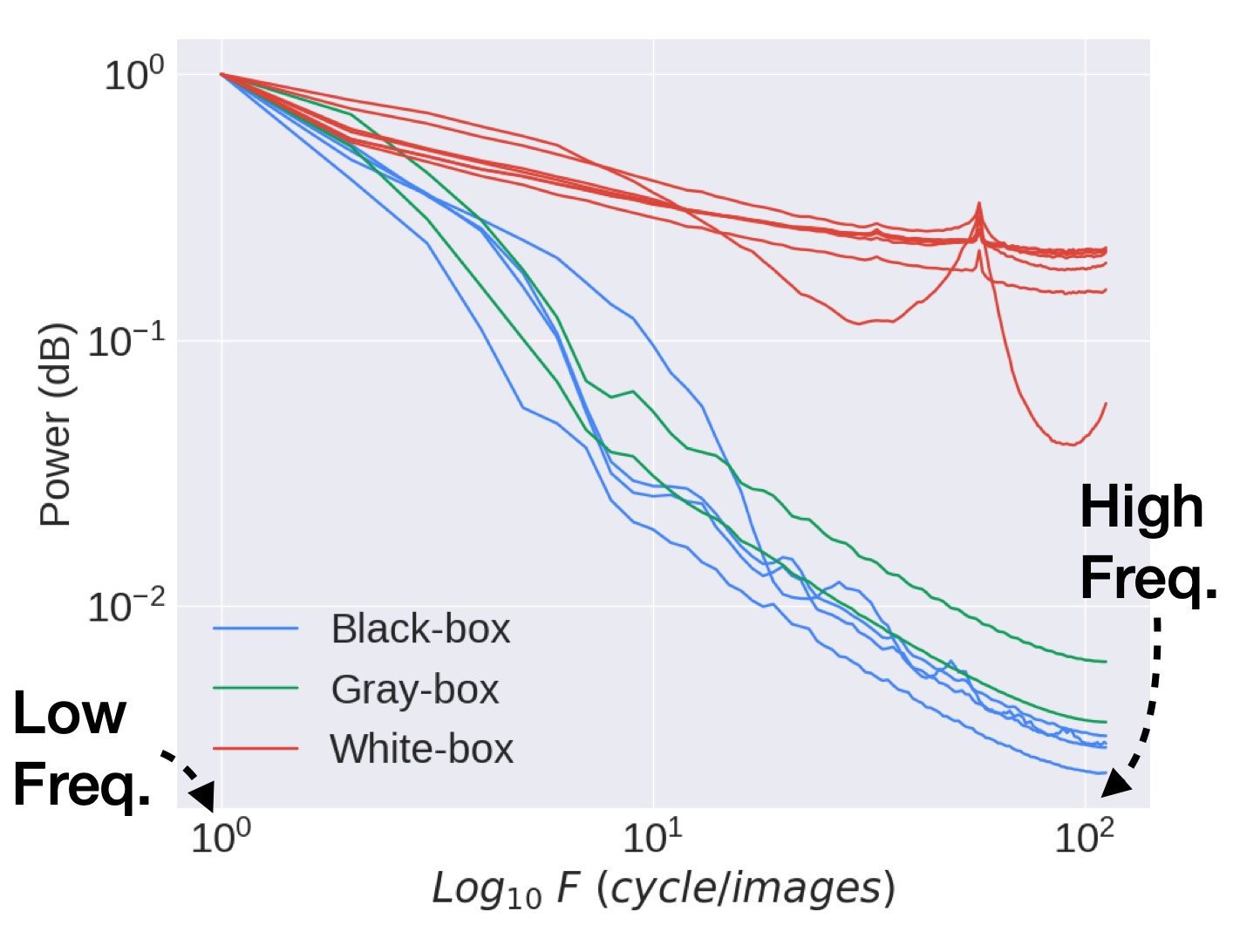}};
\draw [anchor=north west] (0.42\linewidth, 1\linewidth) node {\includegraphics[width=0.58\linewidth]{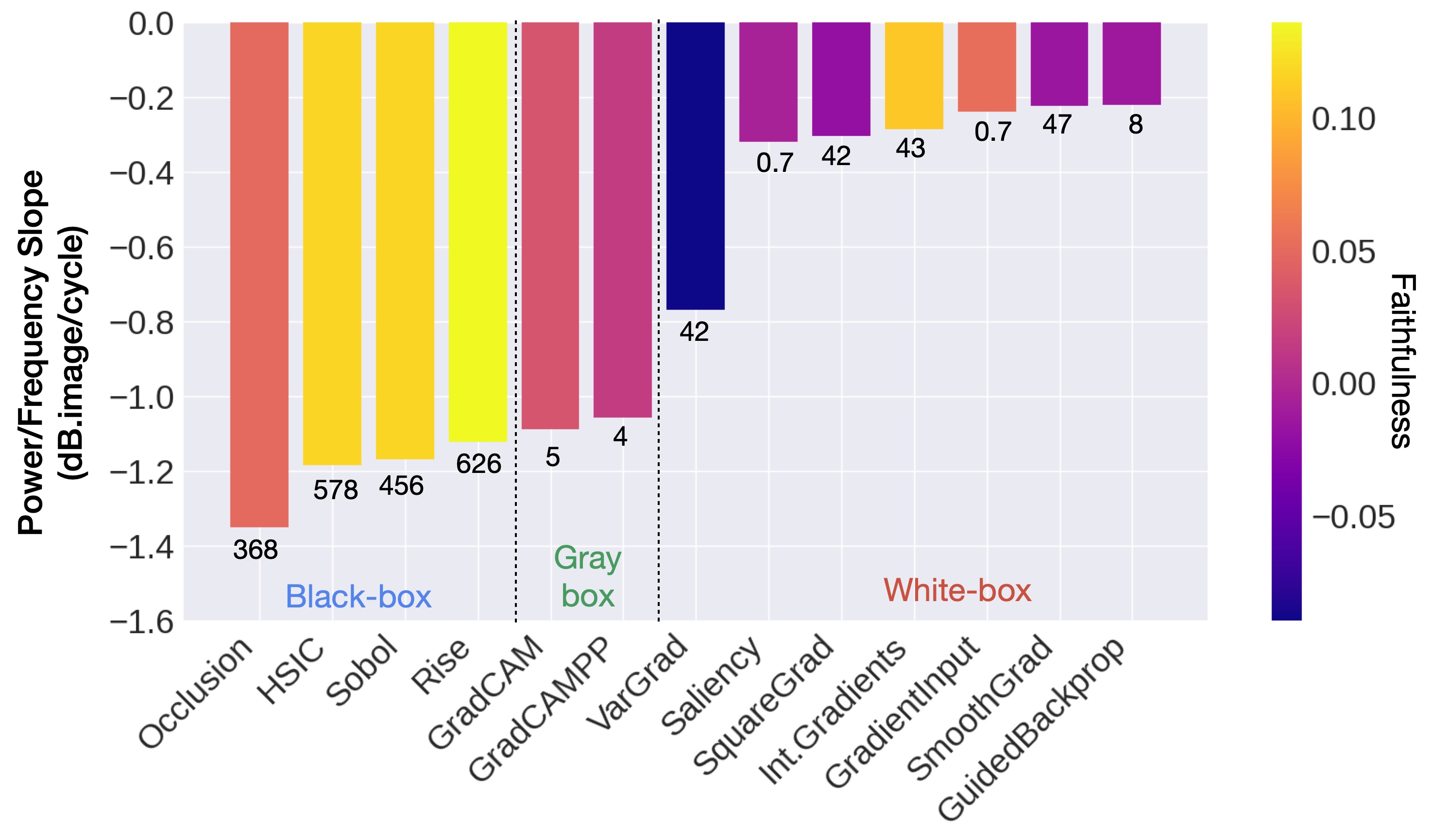}};
\begin{scope}
\draw [anchor=north west,fill=white, align=left] (0.0\linewidth, 1\linewidth) node {\bf A };
\draw [anchor=north west,fill=white, align=left] (0.42\linewidth, 1\linewidth) node {\bf B};
\end{scope}
\end{tikzpicture}
    \caption{\textbf{High-frequency power in attribution methods.} {\bf A:} Fourier signature of white-box and black-box attribution methods. White-box methods produce attribution maps with increased power in the high frequencies.  {\bf B:} Faithfulness and power/frequency slope for several attribution methods. The numbers below the bar correspond to this method's computational time, measured using a set of 100 images (ImageNet) with an Nvidia T4. White-box methods show lower faithfulness but significantly better computational complexity compared to black-box methods. See section~\ref{sec:metrics} for more details on how faithfulness, the power/frequency slope, and the Fourier signature were computed.}
    \label{fig:barchart}
\end{figure*}

\subsection{Comparison between White-Box and Black-Box Methods}

\paragraph{White-box methods are contaminated with high-frequency signal:} Fig.~\ref{fig:barchart}A shows the average Fourier signature for white-box and black-box attribution methods (see Fig.~\ref{fig_app:spectrums2D} for the 2D visualization). We observe that white-box methods produce attribution maps with a greater high-frequency power compared to black-box attribution maps (red curves vs. blue curves in Fig.~\ref{fig:barchart}A). The gray-box methods lie in between white-box and black-box methods.

\textbf{White-box methods are computationally more efficient but have lower faithfulness:} We quantify the computational time of each method to generate an explanation on 100 ImageNet images with an Nvidia T4. The computational time is reported for each attribution method using numbers placed under their corresponding bar in Fig.~\ref{fig:barchart}B. We refer the reader to section~\ref{sec:metrics} for more details on the faithfulness computation. Overall, Fig.~\ref{fig:barchart}B demonstrates that white-box (and gray-box) methods are computationally more efficient than black-box methods. In addition, it shows that white-box methods have a lower power/frequency slope and a lower faithfulness than black-box methods. This suggests that the presence of high-frequency artifacts in white-box methods might be a limiting factor in achieving XAI performance similar to black-box methods.

\subsection{A Low-Pass Filtered Gradient Approximates Well the Original Gradient}
White-box methods produce attribution maps by relying on the gradients of the classifier w.r.t. the input, denoted $\nabla_{\x}\pred(\x)$. The prevalence of high frequencies in these attribution maps suggests that the gradients harbor high-frequency artifacts. To study such artifacts in the gradients, we leverage the first-order approximation of the model, i.e., $\pred(\x+\bm{\ep})\approx \pred(\x)+\ep\nabla_{\x} \pred(\x)$, and we compute the $\ell_2$ approximation error, over different values of $\ep$.

We characterize the error between the Taylor expansion and the actual function value for a specified cut-off frequency of the gradient, denoted as $\sigma$, through the following definition:
$
\zeta(\vx, \sigma) = 
|| \pred(\x+\ep) - (\pred(\x) + \ep \nabla_\sigma \pred(\x)) ||_2
$.
We then define our metric as $\zeta(\x, \sigma) / \zeta(\x, \sigma_{\text{max}})$, with $\sigma_{\text{max}} = 224$.
Here, $\nabla_\sigma \pred(\x)$ represents the gradient at $\x$ with high frequencies removed, up to the cut-off frequency $\sigma$. We compute the error $\zeta$ for various filtered gradients with a cut-off frequency $\sigma \in \{224, \ldots, 5\}$ (where a high $\sigma$ represents a very minor filtering and low $\sigma$ implies that only the very low frequencies are kept), as well as three control conditions representing the absence of information in the gradient.

\begin{figure*}[t!]
\center
\begin{tikzpicture}
\draw [anchor=north west] (0\linewidth, 1\linewidth) node {\includegraphics[width=0.47\linewidth]{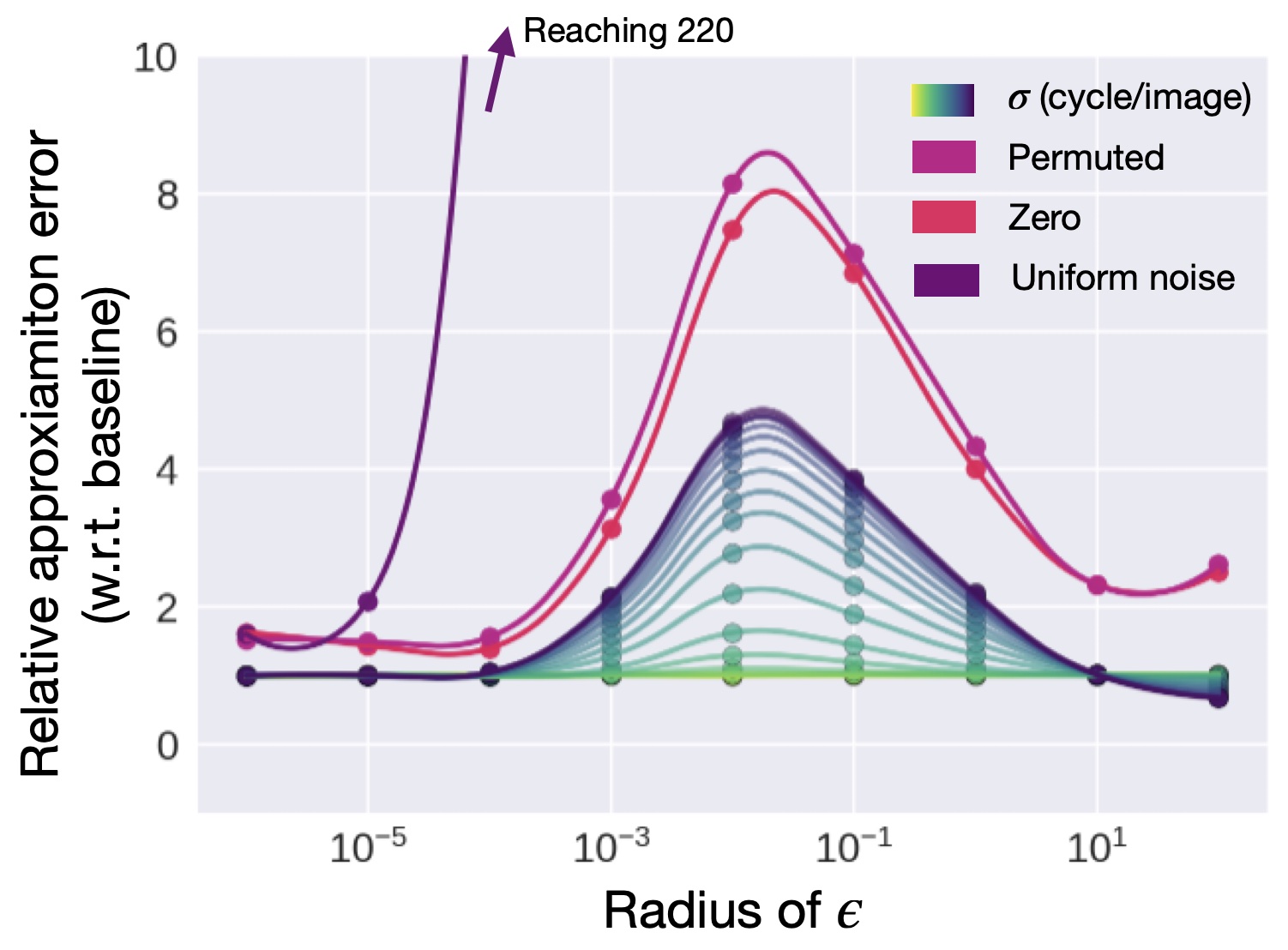}};
\draw [anchor=north west] (0.5\linewidth, 0.98\linewidth) node {\includegraphics[width=0.49\linewidth]{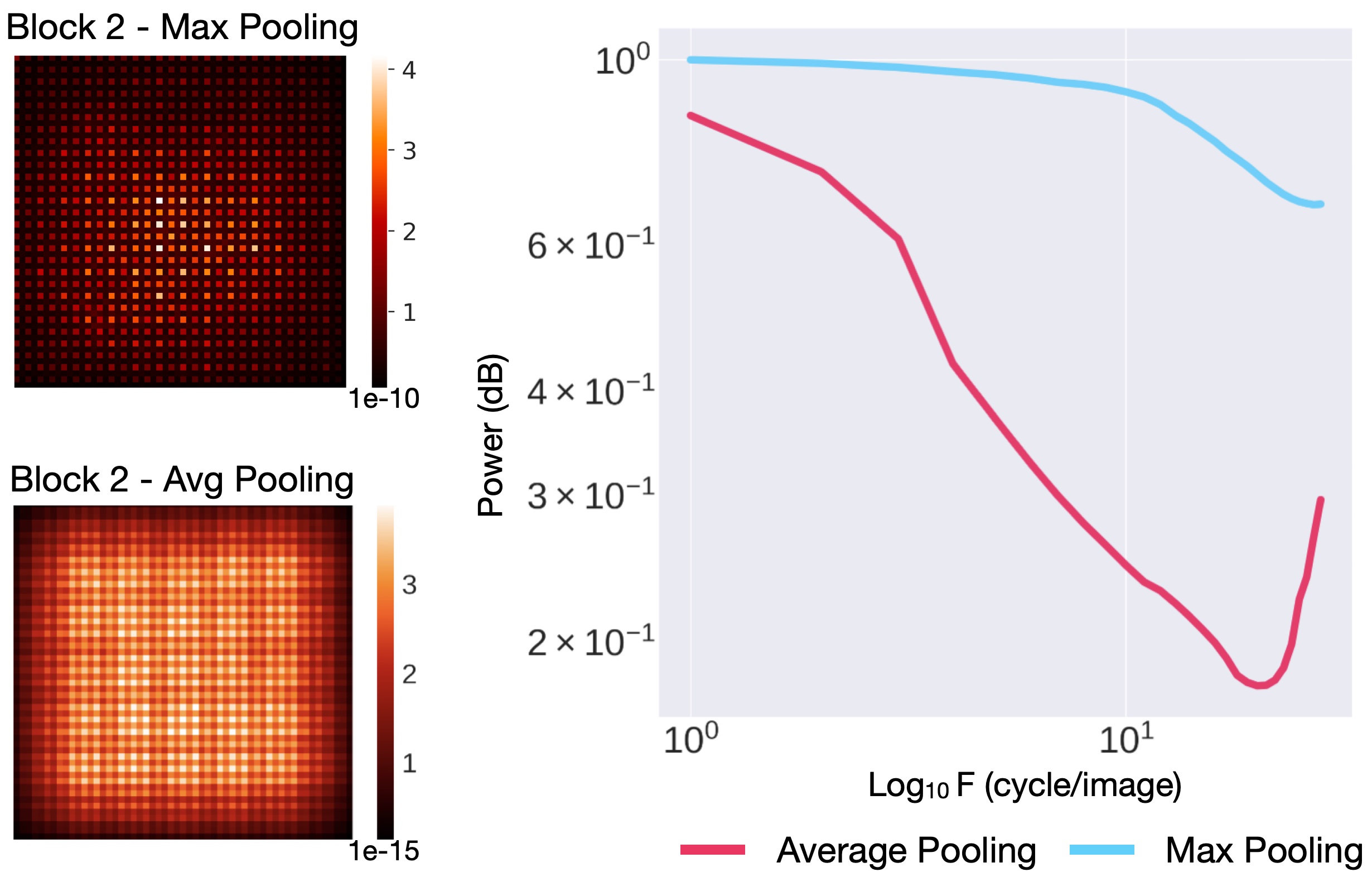}};
\begin{scope}
\draw [anchor=north west,fill=white, align=left] (0.0\linewidth, 1\linewidth) node {\bf A };
\draw [anchor=north west,fill=white, align=left] (0.48\linewidth, 1\linewidth) node {\bf B};
\end{scope}
\end{tikzpicture}
\caption{\textbf{Gradients are contaminated with high frequencies and stem from MaxPooling operations}. {\bf A:} 
We evaluate the importance of high-frequency content in the gradient using a first-order approximation of the model,  i.e., $\pred(\x+\bm{\varepsilon})\approx \pred(\x)+\bm{\varepsilon}\nabla_{\x} \pred(\x)$, 
and we compute the $\ell_2$ approximation error when we remove high frequency up to $\sigma$:  $\zeta(\vx, \sigma)$ relative to the error when we do not filter $\zeta(\vx, \sigma_{\text{max}})$.
Gradients in which we remove high-frequency content (dark blue) produce an error closer to the baseline (green) compared to the control conditions (pink). {\bf B:} Example of gradients in a ResNet architecture after a pooling operation, along with their Fourier signature. MaxPooling operations elicit high-frequency power in the gradient.
}
    \label{fig:uniform_control}
\end{figure*}

In Fig.~\ref{fig:uniform_control}.A, the blueish curves show the evolution of the approximation error from the baseline (gradient not filtered -- most green curve) to the most filtered gradient ($\sigma = 5$ -- most blue curve). We additionally report the approximation for three control conditions (zero gradients, permuted gradients, and uniform noise gradients). In Fig.~\ref{fig:uniform_control}.A, we observe that low-pass filtering gradients elicit a $\ell_2$ approximation error very close to the baseline. These results suggest that the filtered gradient, which remains closer to the first-order term than any of the control conditions, still approximates the non-filtered gradient well when defined as the first-order term of a Taylor expansion. 
Given the linear nature of the first-order approximation, one might consider that the first-order gradient mostly reflects low-frequency content. To control for this, we have included a condition featuring synthetic low-frequency gradients (in the form of zeros everywhere). Such a condition elicits high approximation error (see red curve in Fig.~\ref{fig:uniform_control}.A), suggesting that the first-order approximation of the gradient does not capture only low-frequencies.
We refer the reader to Fig.~\ref{fig_app:appendix_gradient_prediction} for similar results on ConvNeXt and ViT, as well as an additional control condition representing non-informative low-frequency content under the form of a 2D Gaussian.

\subsection{High-Frequency Artifacts Stems from Max-Pooling Operations }
Our investigation now turns to understanding why high-frequency artifacts appear in gradients. In Fig.~\ref{fig:uniform_control}B, we analyze the power spectrum of various gradients in a ResNet50, along with their visualization. Consistent with findings from \citet{olah2017feature}, we observe a clear checkerboard pattern in the gradients following a max-pooling. This pattern is reduced after an average-pooling due to a different mechanism in the backward implementation of both operations (see Fig.~\ref{fig_app:gradients_resnet} for more visualizations). Their respective Fourier signature (Fig.~\ref{fig:uniform_control}B right) confirm our hypothesis: the max-pooling operation shows more power in the high frequencies compared to the average-pooling (see Fig.~\ref{fig_app:resnet_max_avg} for all pooling layers). We additionally investigate the contribution of every layer in the frequency power of the gradients. We analyze the power/frequency slope of gradients across various layers of ResNet50 (see Fig.~\ref{fig_app:resnet_layers}). We observe a jump in the power/frequency slope at each max-pooling operation in between ResNet blocks (see blue curve in Fig.~\ref{fig_app:resnet_layers}). To discern whether this increase in high frequency is a result of the network architecture or the training process, we conduct a similar analysis with an untrained ResNet50 (black curve in Fig.\ref{fig_app:resnet_layers}) and observe a similar trend. Subsequently, we replace max-pooling with average-pooling to ensure the preservation of information continuity in the gradient (red curve in Fig.\ref{fig_app:resnet_layers}). We observe a clear divergence in the power/frequency slope between max-pooling and average-pooling networks, highlighting the crucial role of max-pooling in creating high-frequency signals in gradients. We observe a similar pattern of results with a VGG16~\citep{simonyan2014very} architecture (see Figures~\ref{fig_app:vgg_layers},\ref{fig_app:vgg_max_avg}, \ref{fig_app:gradients_vgg}). Additionally, as noted by \citet{odena2016deconvolution}, this effect can be mitigated by adjusting the stride value during downsampling. Consequently, in Fig.~\ref{fig_app:strides_vgg}, we examine the impact of striding on high-frequency power in a VGG16 network.

In summary, we have established that the high-frequency artifacts contaminating white-box methods originate from the gradient. Specifically, max-pooling operations within the network appear to be a primary source of these high-frequency artifacts. These insights underscore the potential benefits of implementing a low-pass filter with a cut-off frequency adapted to the network architecture to repair white-box attribution methods.

\section{\forgrad: FOurier Reparation of the Gradients}

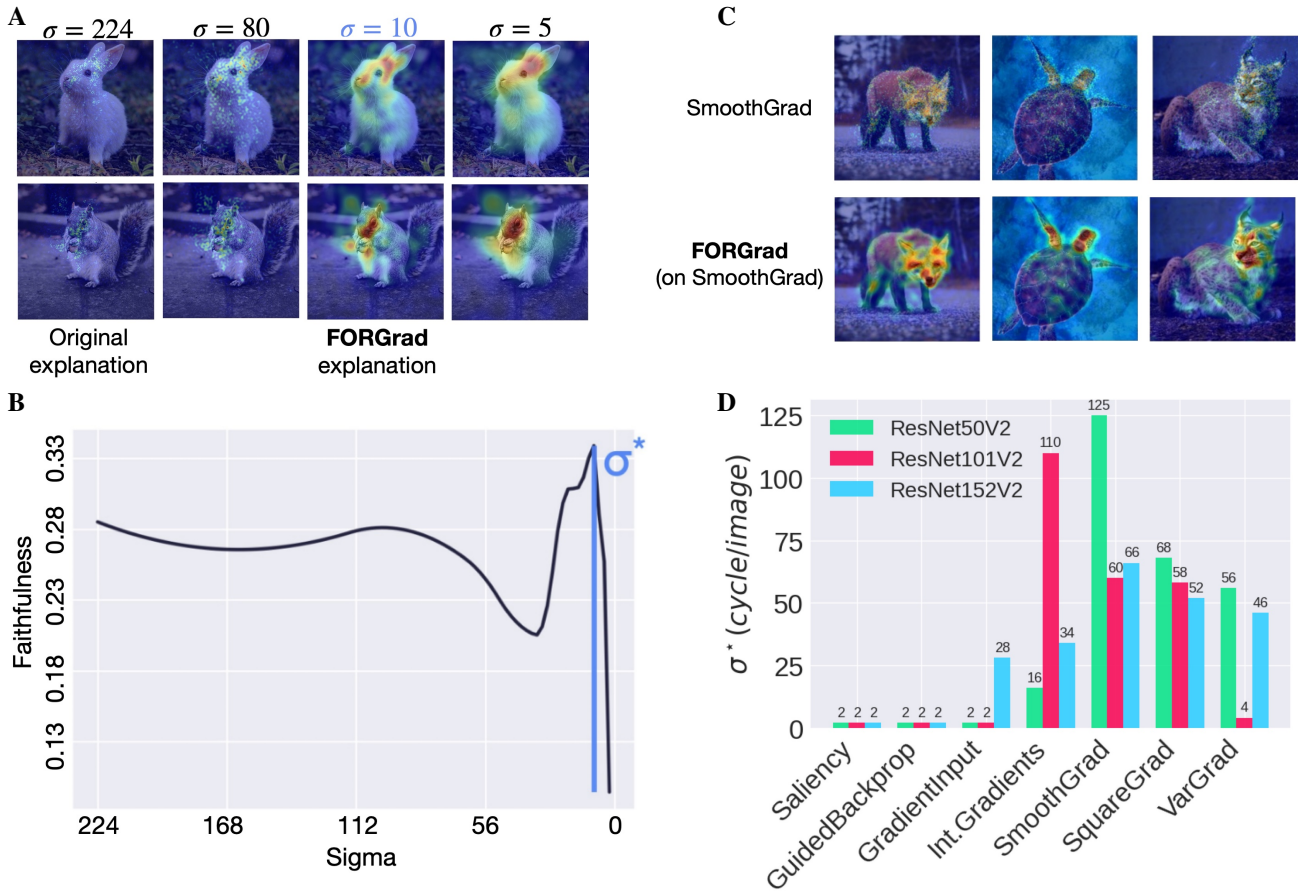
\begin{figure*}[h!]
\center
\center
\begin{tikzpicture}
\draw [anchor=north west] (0\linewidth, 1\linewidth) node {\includegraphics[width=0.45\linewidth]{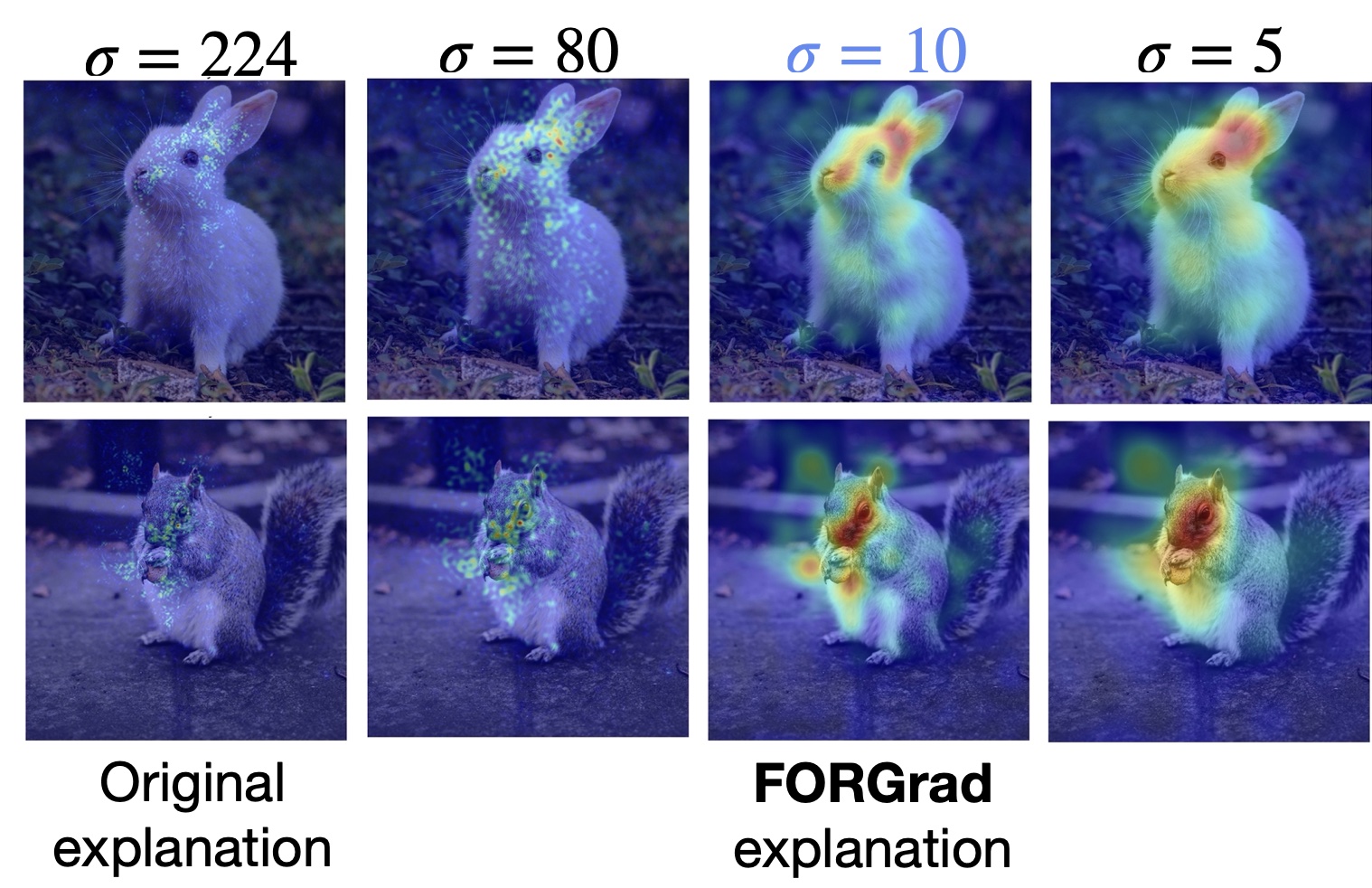}};
\draw [anchor=north west] (0\linewidth, 0.68\linewidth) node {\includegraphics[width=0.50\linewidth]{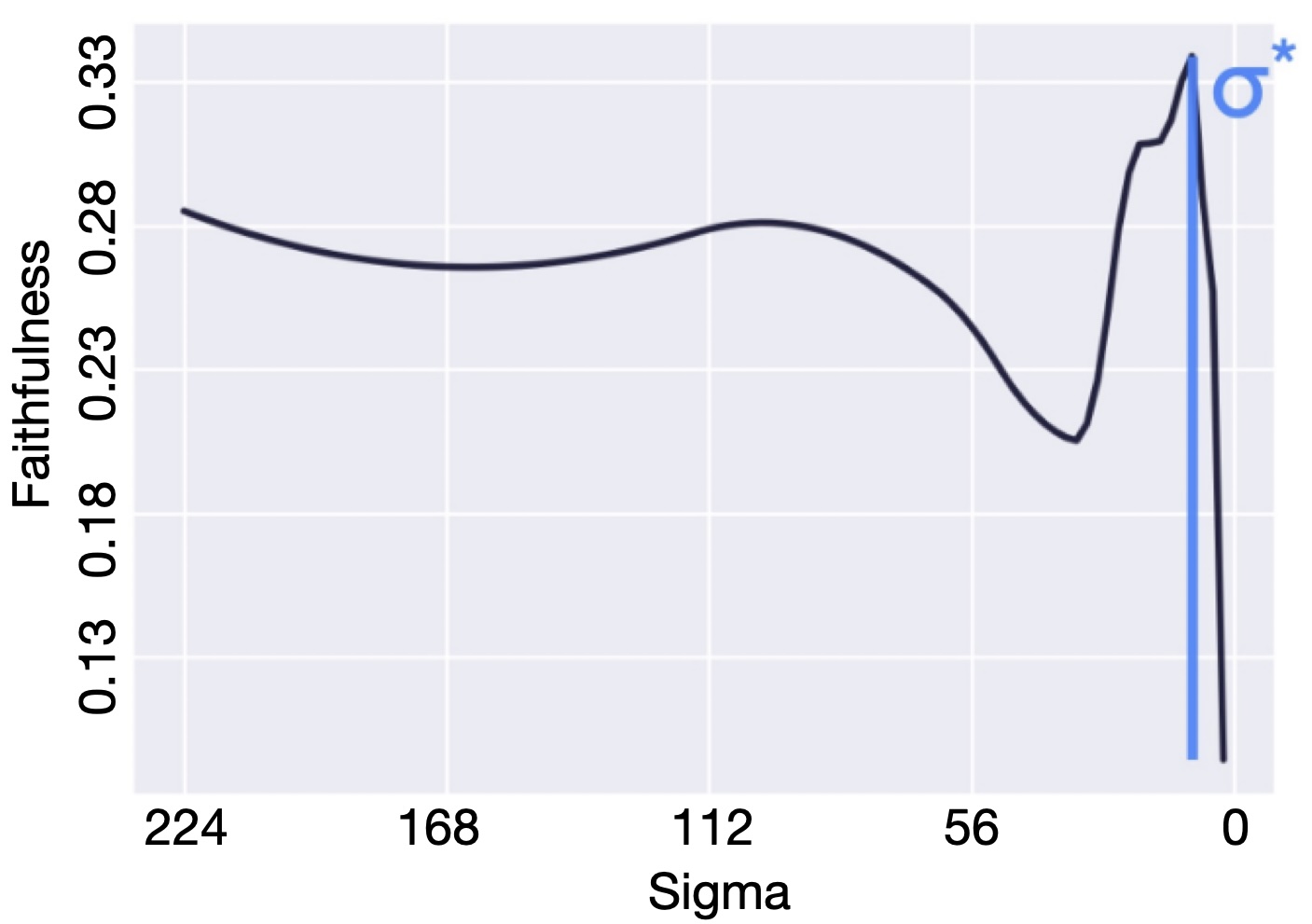}};
\draw [anchor=north west] (0.5\linewidth, 0.98\linewidth) node {\includegraphics[width=0.5\linewidth]{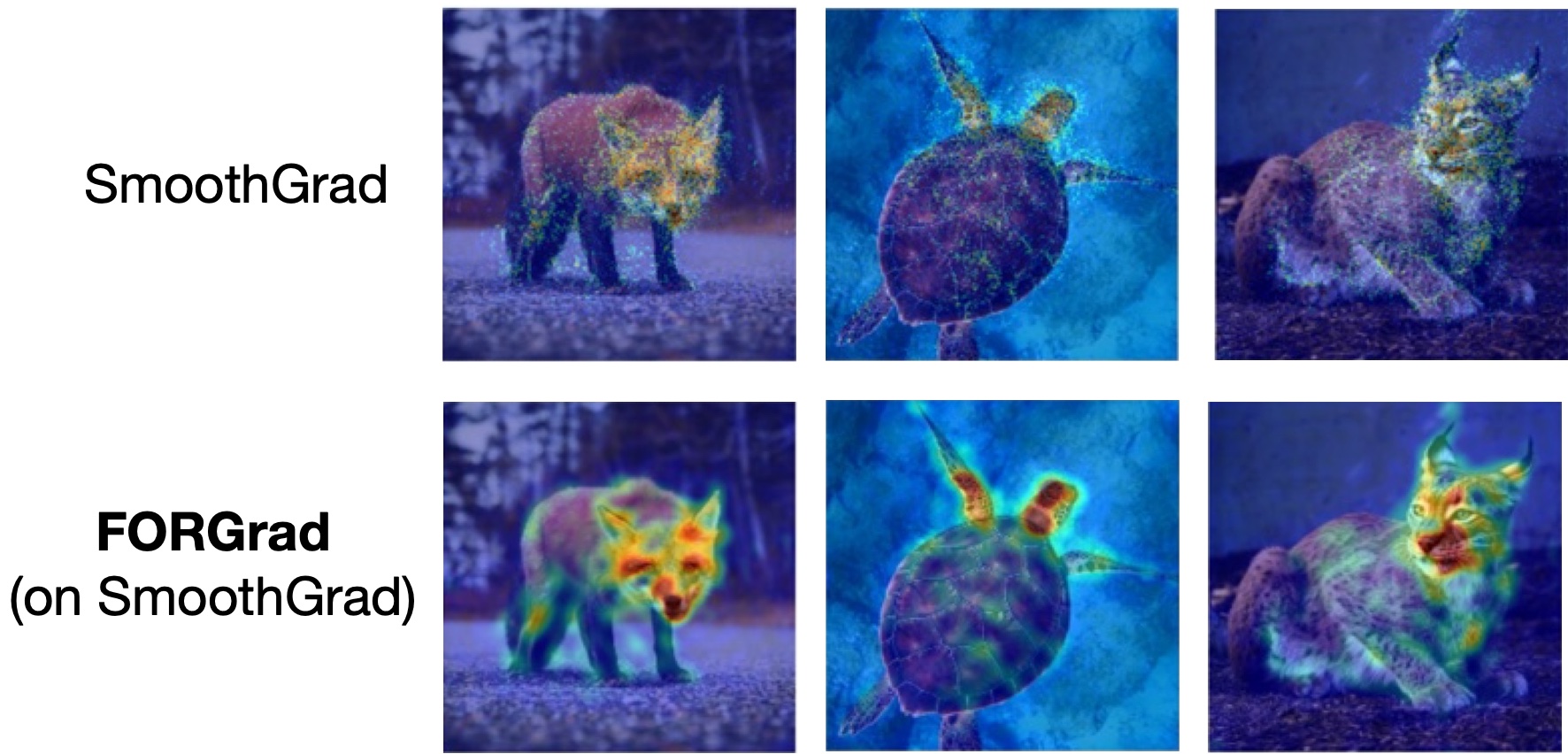}};
\draw [anchor=north west] (0.55\linewidth, 0.70\linewidth) node {\includegraphics[width=0.45\linewidth]{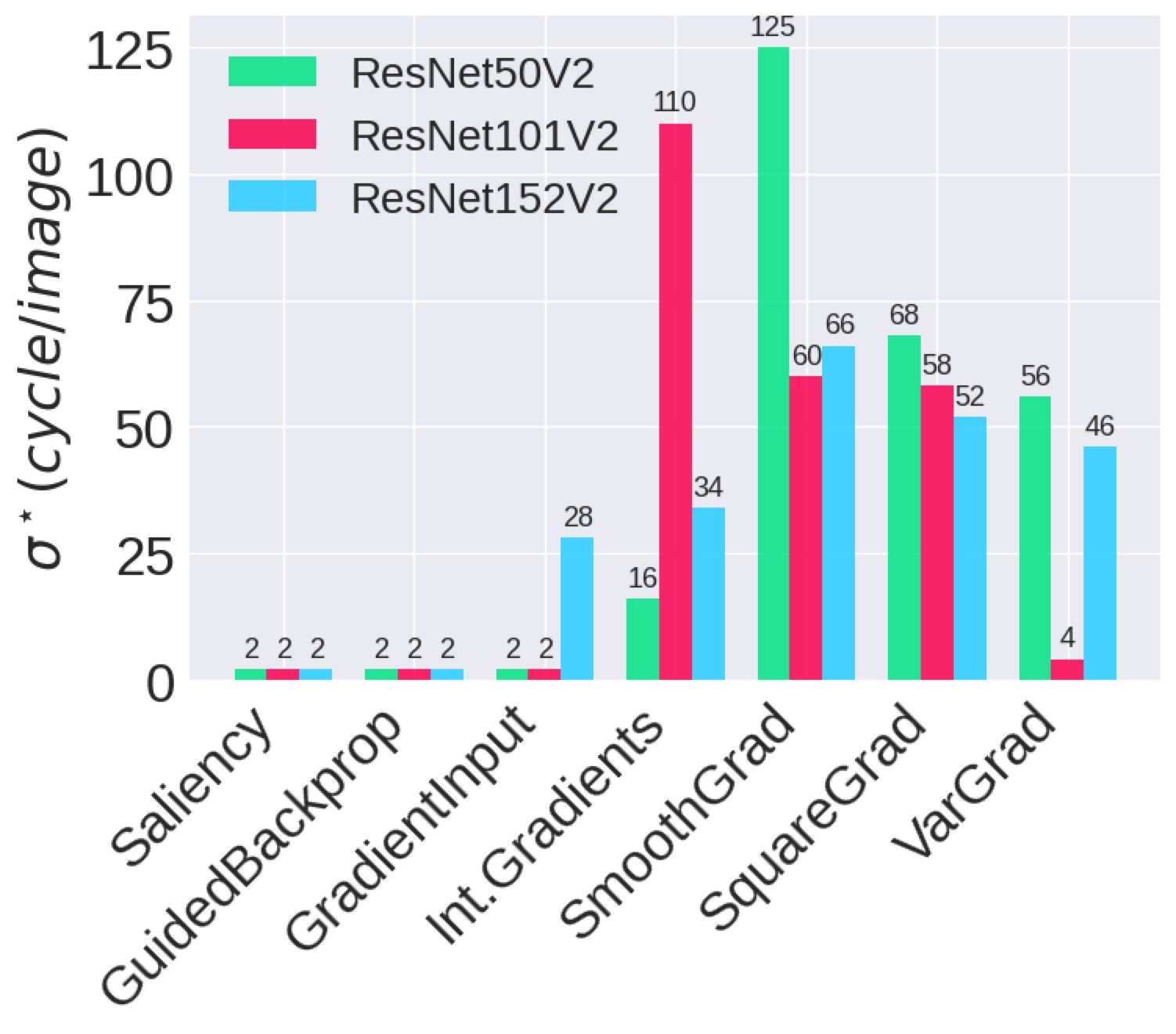}};
\begin{scope}
\draw [anchor=north west,fill=white, align=left] (0.0\linewidth, 1\linewidth) node {\bf A };
\draw [anchor=north west,fill=white, align=left] (0\linewidth, 0.70\linewidth) node {\bf B};
\draw [anchor=north west,fill=white, align=left] (0.55\linewidth, 1\linewidth) node {\bf C};
\draw [anchor=north west,fill=white, align=left] (0.55\linewidth, 0.70\linewidth) node {\bf D};
\end{scope}
\end{tikzpicture}\vspace{-5mm}
\caption{\textbf{The \forgrad~method}. \textbf{A:} Visual examples in which the gradient is low-pass filtered at various cut-off frequencies. Without any filtering ($\sigma=224$), the attribution maps are similar to those obtained with original white-box methods (i.e., Gradient-Input here). \forgrad~finds an optimal cut-off frequency at $\sigma=10$. \textbf{B:} Sensitivity of the faithfulness to
changes in the cut-off frequency. The faithfulness
is evaluated on 1,280 images from the ImageNet validation
set (see section~\ref{sec:metrics} for more details). The frequencies range from $224$, indicative of an unfiltered gradient, to $0$, representing a fully filtered gradient. \textbf{C:} Side-by-side comparison of attribution maps generated using the
SmoothGrad method (top-row) versus
those refined with the \forgrad~method (bottom row). \textbf{D:} Evolution of the optimal $\sigma^{\star}$ values across different ResNetV2 architectures and attribution methods. The variability highlights
the crucial role of a tailored cut-off frequency.}
\label{fig:sigma_selection}
\end{figure*}

With \forgrad~ we propose to filter out high-frequency content that is considered noise to improve the XAI performance of white-box attribution methods. We therefore apply a low-pass filter on the Fourier spectrum of the gradient of the classifier w.r.t the input (i.e. $\nabla_{\x}\pred(\x)$). This low-pass filter is equipped with a mechanism to adjust the cut-off frequency, $\sigma^{\star}$, based on the specific network architecture and attribution method.  

\textbf{An adapted $\sigma^\star$ per model and attribution method:} To identify the ideal cut-off frequency, $\sigma^\star$, we introduce a heuristic to maximize attribution faithfulness. For each network and attribution method, we first assess the relationship between faithfulness and different cut-off frequencies, $\sigma$ in cycle/image, as shown in Fig.\ref{fig:sigma_selection}B (see Fig.~\ref{fig_app:sigmas_models_methods} for the sigma selection per model and method). The faithfulness is evaluated on $1,280$ images from the ImageNet validation set (denoted $\mathcal{V}$) for a spectrum of cut-off frequencies. These frequencies range from $224$, indicative of an unfiltered gradient, to $0$, representing a fully filtered gradient. Detailed information about the faithfulness metric can be found in section~\ref{sec:metrics}. Here, we select the optimal bandwidth $\sigma^{\star}$ such that $\sigma^{\star} = \argmax_{\sigma}\mathbb{E}_{\x\sim\mathcal{V}}F(\am_\sigma(\x))$. 

Fig.~\ref{fig:sigma_selection}B illustrates the sensitivity of the faithfulness to changes in the cut-off frequency, $\sigma$, underscoring the necessity of selecting an optimal $\sigma^{\star}$. In Fig.~\ref{fig:sigma_selection}A, we provide visual examples using the Gradient Input method~\cite{ancona2017better}, where the gradient is low-pass filtered at various cut-off frequencies. One can observe that the initial noisy patterns (Fig.~\ref{fig:sigma_selection}A, $\sigma=220$ ) progressively evolve into clearer and larger patches (Fig.~\ref{fig:sigma_selection}A, $\sigma=80$), culminating in saliency maps that effectively emphasize the critical features for object recognition (Fig.~\ref{fig:sigma_selection}A, $\sigma=10$). However, at very low $\sigma$ values (Fig.~\ref{fig:sigma_selection}A, $\sigma=5$), the attribution map becomes overly diffuse, reducing interpretability. Fig.~\ref{fig:sigma_selection}D compares the optimal $\sigma^\star$ values across different ResNetV2~\cite{he2016identity} architectures and attribution methods. We observe that even with the same attribution method, distinct architectures achieve their maximum faithfulness at different $\sigma^\star$ values. Likewise, the same network architecture can yield different optimal $\sigma^\star$ values when leveraging different attribution methods. This variability highlights the crucial role of a specific cut-off frequency, emphasizing that a one-size-fits-all approach may not be effective in maximizing the faithfulness of attribution methods in different contexts. Additionally, Fig.~\ref{fig:sigma_selection}C presents a side-by-side comparison of attribution maps generated using the SmoothGrad method~\cite{smilkov2017smoothgrad} (top-row) versus those refined with the \forgrad~method (bottom row).  A qualitative analysis shows that the \forgrad~method yields more interpretable explanations, with a sharper focus on crucial features for object recognition. We refer the reader to Fig.~\ref{fig_app:qualitative} for more examples of all white-box methods. 

\begin{table}[t]
    \centering
    \scalebox{0.7}{ %
        \begin{tabular}{p{1mm} p{57mm} p{10mm}p{10mm}p{10mm}p{8mm}}
        \toprule
        && \multicolumn{4}{c}{ResNet50} \\
        \cmidrule(lr){3-6} %
        && Faith.($\uparrow$) & $\mu$Fid.($\uparrow$) & Stab.($\downarrow$) & Time($\downarrow$)
        \\
        \midrule
        \parbox[t]{2mm}{\multirow{16}{*}{\rotatebox[origin=c]{90}{White-box}}}& Saliency\cite{simonyan2013deep} &   
                  0.18 & 0.40 & 0.67 & 0.78 \\ 
        & Saliency$^\star$ &   
                  0.28 & 0.39 & 0.53 & 0.89 \\ 
        \cmidrule(lr){2-6}
        & Guidedbackprop\cite{ancona2017better} &
                  0.31 & 0.45 & 0.28 & 8.25 \\
        & Guidedbackprop$\star$ &
                  0.35 & 0.45 & 0.22 & 7.05 \\
        \cmidrule(lr){2-6}
        & GradInput\cite{shrikumar2017learning} &
                  0.2 & 0.36 & 0.42 & {\bf 0.73} \\ 
        & GradInput$\star$ &
                  0.26 & 0.36 & 0.35 & \underline{0.77} \\ 
        \cmidrule(lr){2-6}
        & Int.Grad\cite{sundararajan2017axiomatic} &   
                  0.24 & 0.39 & 0.72 & 42.7 \\
        & Int.Grad$\star$ &   
                  0.31 & 0.38 & 0.76 & 41.3 \\
        \cmidrule(lr){2-6}
        & SmoothGrad\cite{smilkov2017smoothgrad} &   
                  0.23 & 0.45 & 0.22 & 46.6 \\
        & SmoothGrad$\star$ &   
                  \underline{0.37} & 0.44 & 0.21 & 48.3 \\ %
        \cmidrule(lr){2-6}
        & VarGrad \cite{adebayo2018sanity} &   
                  0.36 & \underline{0.46} & {\bf 0.003} & 41.5 \\
        & VarGrad$\star$ &   
                  0.35 & 0.44 & \underline{0.004} & 40.6 \\ %
        \cmidrule(lr){2-6}
        & SquareGrad\cite{seo2018noise} &   
                  0.36 & 0.45 & {\bf 0.003} & 42.1 \\
        & SquareGrad$\star$ &   
                  0.36 & \underline{0.46} & 0.005  & 40.9 \\ 
        \midrule
        \midrule
        \parbox[t]{2mm}{\multirow{6}{*}{\rotatebox[origin=c]{90}{Black \& Gray-box}}} & GradCAM\cite{Selvaraju_2019} &   
                  0.31 & 0.40 & 0.31 & 5.24 \\
        & GradCAM++\cite{chattopadhay2018grad} &
                  0.33 & 0.43 & 0.34 & 4.61 \\
        & Occlusion\cite{ancona2017better} &   
                  0.20 & 0.39 & 0.6 & 368 \\
        & HSIC\cite{novello2022making}     &   
                  0.33 &  {\bf 0.47} &  0.45 & 456 \\
        & Sobol\cite{fel2021sobol}     &   
                  0.34 &  {\bf 0.47} &  0.47 & 578 \\
        & RISE\cite{petsiuk2018rise}     &   
                  {\bf 0.41}  &  0.34 &  0.55 & 626 \\
        \bottomrule \\
            \end{tabular}
    }
    \caption{{\bf Results on Faithfulness and Stability metrics}. Faithfulness, Fidelity, and Stability scores were obtained on 1,280 ImageNet validation set images, on an Nvidia T4 (For Faithfulness and Fidelity, higher is better).
    Time in seconds corresponds to the generation of 100 (ImageNet) explanations on an Nvidia T4.
    The first and second best results are in \textbf{bold} and \underline{underlined}.}%
    \label{tab:fathifulness}
    \vspace{-5mm}
\end{table}
\textbf{\forgrad~allows white-box methods to rival with black-box methods:}
We now evaluate the \forgrad~method using XAI benchmarks, focusing on three established metrics: Faithfulness, $\mu$Fidelity, and Sensitivity (detailed in section ~\ref{sec:metrics}). Our findings, presented in Table~\ref{tab:fathifulness}, compare the original performance of white-box attribution methods, their enhanced versions via the \forgrad~method (marked with a $\star$), black-box, and gray-box methods. This comparison also includes the computational time measured on 100 images with an Nvidia T4 required to produce attribution maps. We showcase results on a ResNet50 network~\cite{he2016deep} in Table~\ref{tab:fathifulness}, and provide additional results for ConvNext~\cite{Liu_2022_CVPR} and ViT~\cite{dosovitskiy2020image} in Table \ref{tab:fathifulness_full}. All metrics are computed on a subset of the ImageNet validation set made of $1280$ images. Importantly, this subset does not include the images used to compute $\sigma^{\star}$. Three key observations emerge from Table~\ref{tab:fathifulness}. First, \forgrad~consistently boosts the faithfulness metrics for white-box methods (the quantitative improvement is validated by a statistical effect given by a Bayesian Anova analysis -- section \ref{stats}). This observation experimentally validates the \forgrad~method. Second, the $\mu$Fidelity metric shows little variations for most of the \forgrad~white-box methods. We attribute this observation to the fact that the $\mu$Fidelity metric is strongly biased toward high-frequency content: random attribution maps with higher-frequency contents exhibit higher $\mu$Fidelity (see Table \ref{tab:control_random} and Fig.~\ref{fig_app:random_exp}). However, the sensitivity metric, similarly to the faithfulness, is systematically improved for most models and methods, despite being evaluated with a $\sigma^{\star}$ optimized on the faithfulness.
Third, in terms of computations, white-box (and gray-box) methods outperform black-box approaches. This increased efficiency stems from white-box attribution methods using more information per network pass, thereby reducing the need for multiple network queries. 
Additionally, we validate the optimality of our method by computing~\forgrad~on the attribution maps instead of the gradients. In Table~\ref{tab:metrics_std}, we compare the non-filtered metrics with the ones obtained with~\forgrad~on the attribution maps and~\forgrad~on the gradients (before computing the explanation). We observe a systematic improvement of the scores with~\forgrad~but more significant when applied to the gradients.
Finally, we ensure that~\forgrad~does not affect the validity of white-box methods according to \citet{adebayo2018sanity} sanity check. We show in Fig.~\ref{fig_app:sanity_check} the resulting maps after randomizing the weights of a ResNet50 on white-box methods combined with \forgrad. The methods already failing the test before~\forgrad~still do, while the successful ones continue to pass the sanity check. All in all, \forgrad~discards high-frequency artifacts to enhance low-frequency content of the attribution maps provided by the white-box methods by filtering out the noisy high-frequencies in the gradient, making such methods rival with low-frequency black-box methods, as summarized in Fig.~\ref{fig_app:appendix_summary}.  

\begin{table}[t]
    \centering
    \scalebox{0.7}{ %
        \begin{tabular}{l cc}
        \toprule
        & \multicolumn{2}{c}{ResNet50} \\
        
        & Original & \forgrad \\
        
        \midrule
        
         1 & VarGrad (0.82) & \best{SquareGrad$\star$ (0.82)}  \\

         2 & SquareGrad (0.81) & \best{VarGrad$\star$ (0.82)}  \\

         3 & GuidedBackprop (0.48) & \best{SmoothGrad$\star$ (0.60)} \\
         
         4 & SmoothGrad (0.46) & \best{GuidedBackprop$\star$ (0.58)}  \\

         5 & GradCAM++ (0.42) & GradCAM (0.42)  \\

        \bottomrule \\
    \end{tabular}
    }
    \caption{{\bf Global ranking before (original) and after \forgrad.} For each model, we show the five attribution methods with the highest metrics, before and after applying \forgrad. The explanation maps were computed on $1280$ images from the validation set of ImageNet, based on an aggregation of the three metrics computed by $(F(\x, \am) + \mu F(\x, \am) - S(\x, \am)$.
    }
    \label{tab:ranking}
\end{table}
In Table~\ref{tab:ranking}, we aim to find the best attribution methods for ResNet50. To do so, we synthesize the faithfulness, $\mu$fidelity, and sensitivity metrics into one single, comprehensive score. This score, labeled  $(F + \mu F) - S$, corresponds to the sum of the faithfulness and the $\mu$fidelity, minus the sensitivity. Table \ref{tab:ranking} reveals a notable shift in method rankings when integrating \forgrad. This change results in at least $4$ white-box methods in the top-5 position for the ResNet50 network. Similar findings for ConvNext and ViT are detailed in Table~\ref{tab:ranking_full}. 

In conclusion, the \forgrad~method not only enhances the performance of white-box methods to rival or even surpass black-box methods but also achieves this with greater computational efficiency.

\section{Conclusion and Perspectives}
This research started with the observation that white-box and black-box attribution methods produce attribution maps with different power spectra. Specifically, white-box methods show greater high-frequency power. We found that this high-frequency content originates from the gradient information used by white-box methods. We demonstrate that selectively filtering out these high frequencies does not alter the approximation error of the original gradient (using a first-order approximation). It suggests that such high-frequency content is non-informative and can be considered as noise. Further analysis revealed a strong connection between high frequencies in gradients and certain architectural constraints, including max-pooling operations. This led us to hypothesize that filtering out these high frequencies would enhance white-box attribution methods that rely on gradient information. To address this, we developed \forgrad~ a low-pass filter with an optimal cut-off frequency tailored to each model architecture and attribution method. The application of \forgrad~resulted in substantial improvements in XAI performance across all white-box attribution methods, challenging the traditional hierarchy of white-box and black-box methods. Notably, with \forgrad~ white-box methods now dominate the top-5 rankings while also being more computationally efficient than their black-box counterparts.

Our current results demonstrate the efficiency of~\forgrad~on gradients containing artifacts. For this reason, the more high frequency is contained in the gradient, the more striking the improvement with~\forgrad~will be. Conversely, if filtering does not ameliorate the scores,~\forgrad~will not filter out anything, resulting in the original unfiltered explanation. Such a case is likely to happen with networks containing smooth gradients by design, such as 1-Lipschitz networks~\citep{fel2022good}.

Our finding that the max-pooling operations significantly amplify high frequency in the gradient w.r.t input is surprising. This observation opens the doors to future research focused on designing novel pooling techniques that reduce high frequencies in the gradient itself. Consistent with these results, the impact of~\forgrad~on ViT is less striking than on the Convolution model. This implies that a different phenomenon could happen in Transformer architectures and an investigation could lead to a better interpretability of this family of networks.
Additionally, conducting a systematic analysis of the Fourier spectrum of other architectural components, such as the normalization or attention layers, could lead to further improvements. Ideally, such an exploration would result in a comprehensive list of each architectural element's contribution to the decision process. Studying their respective impacts would contribute to building neural networks that are interpretable by design. 

The impact of~\forgrad~is measured by the standard metrics of XAI. We evaluated~\forgrad~on three different metrics and observed a clear improvement in two of them. However, finding appropriate metrics is still an open question in this field, and these results could be affected by the development of new metrics. 
In~\citet{bhatt2020evaluating}, the authors advocate for a way to combine several XAI metrics to obtain the best method. They propose an aggregate of Faithfulness, Sensitivity, and Complexity of the explanation, arguing that the best explanation should maximize the first and minimize the latter two. Here, we evaluate~\forgrad~on two of these metrics and complete this framework with the Computational cost measured by the time in seconds. We, therefore, believe that~\forgrad~remains an optimal solution to compute faithful explanations in a reasonable amount of time, given the current metrics to evaluate the quality of an explanation. 
More interestingly, one possible explanation for why black-box methods tend to outperform white-box methods is that their optimization objectives align with the evaluation metrics (e.g., Occlusion with Deletion metric). In contrast, the gradients result from optimizing the model for a classification task. Therefore, it is encouraging to achieve scores comparable to those of black-box methods using white-box attribution methods with \forgrad~ bringing Saliency and other gradient-based methods back into prominence.

So far, we have focused on gradient filtering to improve white-box attribution methods. One promising avenue would be to filter out high-frequencies in the gradient during training to reduce the presence of noisy artifacts during the optimization process and inherently improve its quality. We hypothesize that this approach may lead to better performance and faster convergence, as well as more qualitative information to explain the model using white-box methods.
Networks equipped with such a filtering mechanism may also demonstrate enhanced adversarial robustness. 
We hope our work will spur interest in studying network frequency profiles, as it could lead to major improvements   in terms of interpretability and robustness. 
Overall, we anticipate that our research will foster broader adoption of simpler and more efficient white-box methods for explainability, offering a better balance between faithfulness and computational efficiency.

\section*{Acknowledgments}
This work was supported by grants from ONR (N00014-19-1-2029) and NSF (IIS-1912280 and EAR-1925481) to TS and RV and the ANR-3IA Artificial and Natural Intelligence Toulouse Institute (ANR-19-PI3A-0004).
The Carney Institute for Brain Science and the Center for Computation and Visualization (CCV) provided additional support. We acknowledge the Cloud TPU hardware resources that Google made available via the TensorFlow Research Cloud (TFRC) program and computing hardware supported by NIH Office of the Director grant S10OD025181.

\section*{Impact Statement}
This paper presents work whose goal is to advance the field of Machine Learning. There are many potential societal consequences of our work, none of which we feel must be specifically highlighted here.

\bibliography{example_paper}
\bibliographystyle{icml2024}

\newpage
\appendix
\onecolumn
\counterwithin{figure}{section}
\section{Appendix}

\subsection{Additional results}

\begin{figure*}[h]
\center
\includegraphics[width=0.99\textwidth]{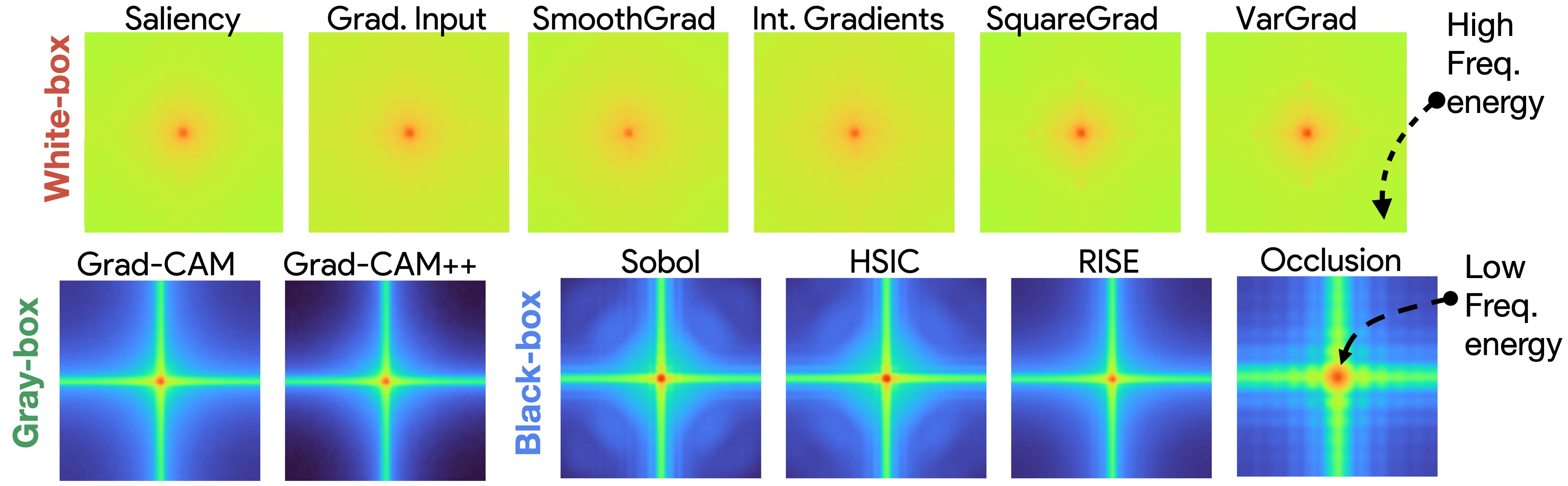}
\caption{\textbf{2D Fourier spectrum of attribution methods.} We show on the top row the Fourier spectrum of white-box attribution methods and of the gray and black-box methods on the bottom row, computed with a ResNet50V2 (each spectrum is normalized to highlight values ranging within the same boundaries). The three families can be distinguished by methods but also by their signature in the Fourier domain. The former method has magnitudes largely concentrated in the low frequencies, while the latter is more spread out: it features non-trivial magnitudes almost everywhere, including in high frequencies. (Blue = low power, Green = high power, Red = very high power)}
\label{fig_app:spectrums2D}
\end{figure*}

\paragraph{Frequency power per category.} Our quantitative measure of high-frequency power derives from the Fourier signature (the 1D Fourier spectrum). The spectra per method, as shown in Fig.~\ref{fig:barchart}, show a clear difference between categories. As these spectra are derived from their 2D equivalent, we provide in Fig.~\ref{fig_app:spectrums2D}, the 2D visualization. These spectra highlight the same information: white-box methods show a Fourier spectrum with a lot of power in the high-frequency, conversely to gray and black-box methods. 

\newpage
\subsection{Gradient approximation with ConvNeXt and ViT}
In this section, we show the results of the gradient approximation presented in the main text on ConvNeXt and ViT. For each network, we evaluate the importance of high-frequency content in the gradient using a first-order approximation of the model,  i.e. $\pred(\x+\bm{\varepsilon})\approx \pred(\x)+\bm{\varepsilon}\nabla_{\x} \pred(\x)$, and we compute the $\ell_2$ approximation error, i.e. $\lVert\pred(\x+\bm{\varepsilon})- (\pred(\x)-\epsilon.\nabla_{\sigma})\rVert_{2}/\lVert\pred(\x+\bm{\varepsilon})- (\pred(\x)-\epsilon.\nabla_{\sigma_{max}})\rVert_{2}$. We additionally provide four control conditions representing the absence of relevant information in the gradient using a gradient containing permuted content, zeros, uniform noise, or random 2D gaussians. The left plot shows the results on ConvNeXt and the right plot for ViT. The reader can notice that compared to Fig.~\ref{fig:uniform_control}.A, the approximation errors of the permuted and zero control conditions reach an order of magnitude more than the most filtered gradient, confirming our precedent results with ResNet50.
The 2D Gaussian control condition represents information of low-frequency content under the form of randomly located 2D Gaussians instead of the gradient. Contrary to the ``zero'' condition (also representing low-frequency information), the activity is not null and might be more informative. However, it reaches the same approximation as the other baseline, being at least twice as high as the more filtered gradient. These additional results confirm the hypothesis that high frequencies in the gradient can be considered not as informative as the low frequencies.

\begin{figure*}[h]
\center
\includegraphics[width=0.99\textwidth]{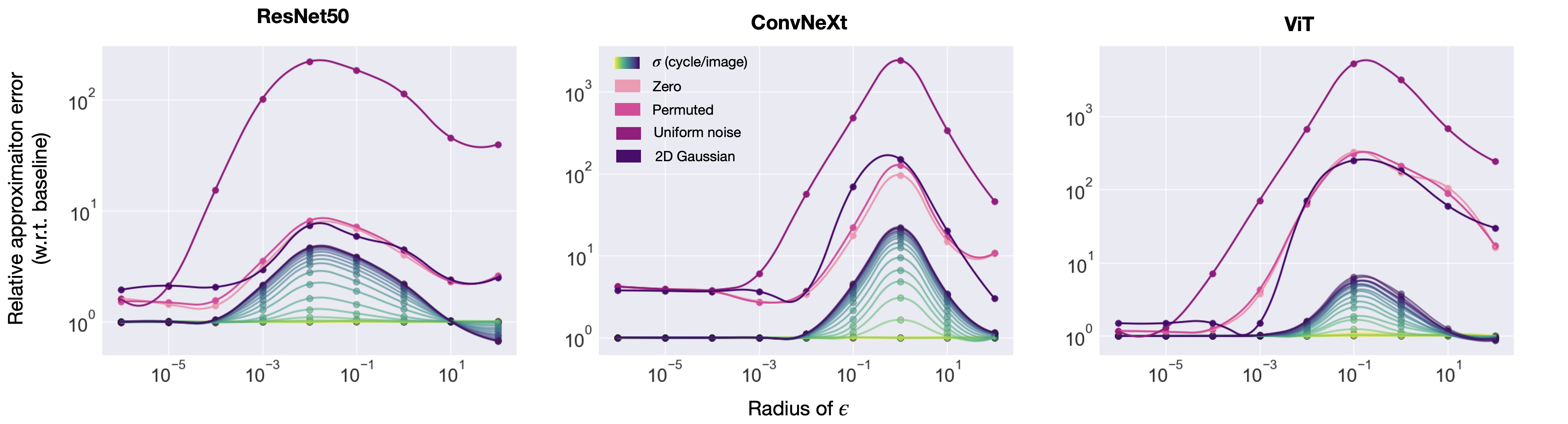}
\caption{{\bf Gradient approximation experiment on ResNet50, ConvNeXt and ViT.} We evaluate the importance of high-frequency content in the gradient using a first-order approximation of the model,  i.e. $\pred(\x+\bm{\varepsilon})\approx \pred(\x)+\bm{\varepsilon}\nabla_{\x} \pred(\x)$, and we compute the $\ell_2$ approximation error, i.e. $\lVert\pred(\x+\bm{\varepsilon})- (\pred(\x)-\epsilon.\nabla_{\sigma})\rVert_{2}/\lVert\pred(\x+\bm{\varepsilon})- (\pred(\x)-\epsilon.\nabla_{\sigma_{max}})\rVert_{2}$. Gradients in which we remove high-frequency content (dark blue) produce an error closer to the baseline (green), compared to the control conditions (purple).}
\label{fig_app:appendix_gradient_prediction}
\end{figure*}

\newpage
\subsection{Investigating the impact of MaxPooling on high-frequency content}
The following section details the role of MaxPooling in introducing high-frequency artifacts in the gradient. We obtain the gradient w.r.t the input by running an inference on a model (pre-trained or not), and back-propagating the information back to the input layer. In the section, we are therefore interested in the direction of the information from the prediction to the input, and we investigate the role of each operation on the content of the gradient.

\begin{figure*}[h]
\center
\includegraphics[width=0.99\textwidth]{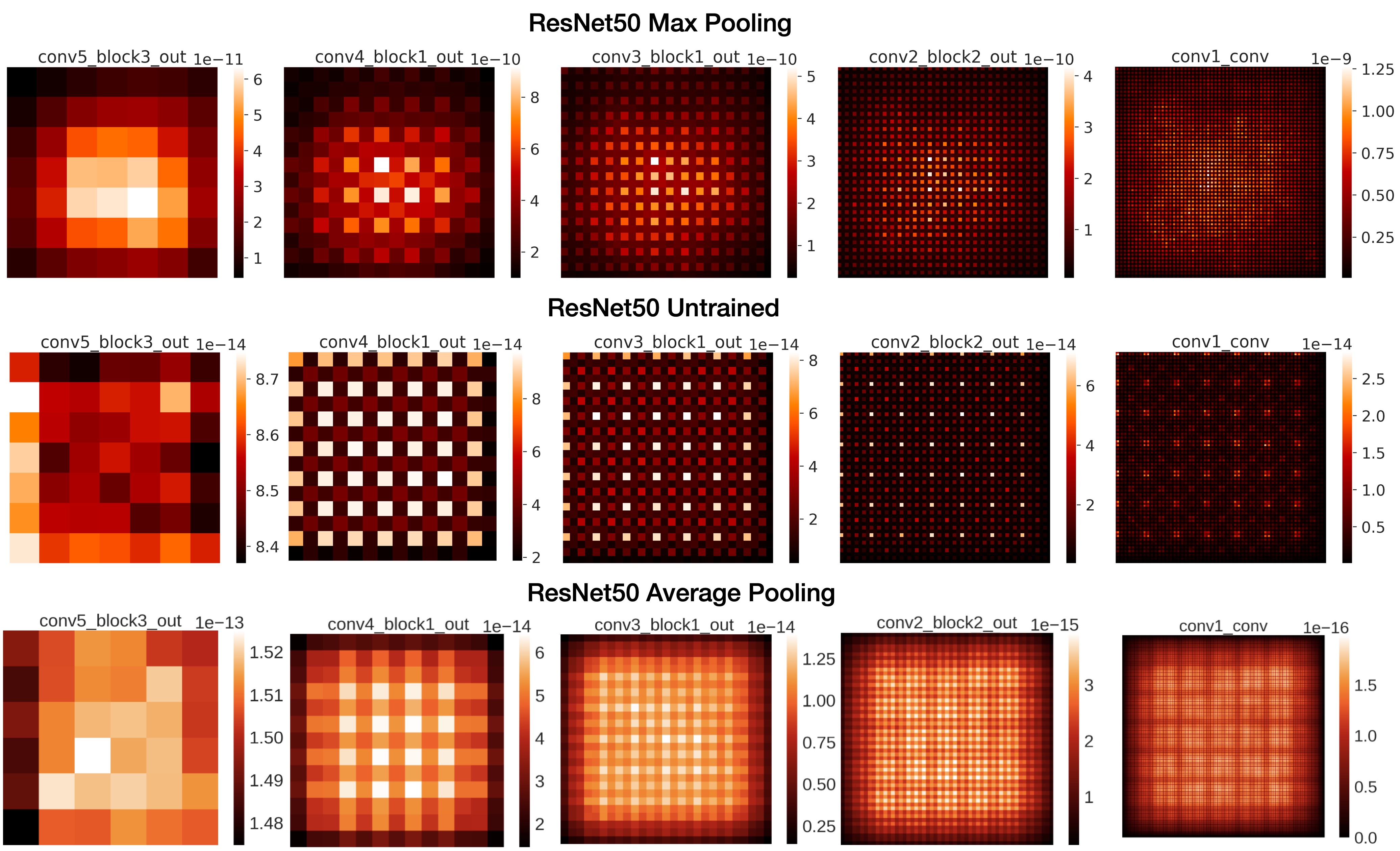}
\caption{Visualization of the gradient (w.r.t. the input) probed at different layers. We observe that both in the ResNet MaxPooling and the ResNet untrained, a checkerboard signal emerges in the gradients. Such a signal might cause an increased power in high frequencies. In contrast, this signal is attenuated in the ResNet50 Average Pooling. It suggests that MaxPooling operations, creating such a checkerboard pattern, might contaminate the gradients with high-frequency noise.}
\label{fig_app:gradients_resnet}
\end{figure*}

\paragraph{Downsampling operations in a ResNet50V2.}
We start by providing additional visualizations of the gradient at several stages of a ResNet50V2 in Fig.~\ref{fig_app:gradients_resnet}. We compare the content of a gradient provided by a pre-trained model on ImageNet, a random model, and a model where the MaxPooling operations as well as started convolutions were replaced by an AveragePooling.
The first block, starting from the prediction -- and denoted \textit{conv5} here -- shows no strong high-frequency artifact. However, after applying a downsampling operation (upsampling on the gradient), we observe some checkerboard-like patterns emerging in the architectures with MaxPooling. The pattern is however minimized in the architecture doted of AveragePooling. The deeper we look into the network, the more high frequencies are present in the gradient.

\begin{figure*}[h]
\center
\includegraphics[width=0.5\textwidth]{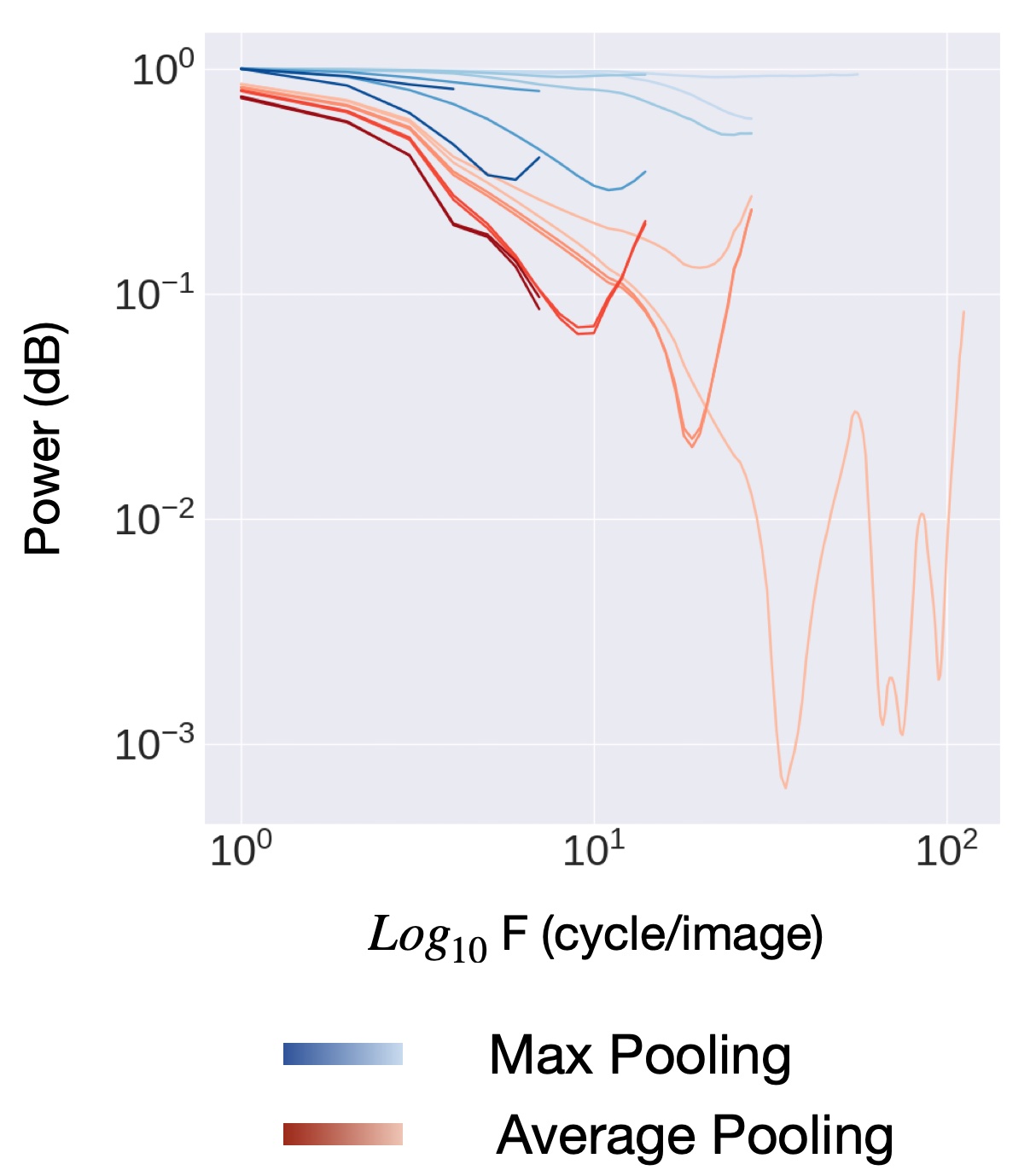}
\caption{{\bf Fourier signature of the gradient after pooling operations in a ResNet50.} The gradients of a ResNet50 after MaxPooling operations (blueish curves) show a Fourier signature with more power in the high-frequencies compared to after an AveragePooling (reddish curves).}
\label{fig_app:resnet_max_avg}
\end{figure*}

To confirm quantitatively this observation, we compute the Fourier signature of these gradients, quantifying the frequency power at each block. The resulting curves are shown in Fig.~\ref{fig_app:resnet_max_avg}. The gradients coming from a ResNet50 with MaxPooling (blue curves) contain all more power in the high-frequencies, compared to gradients issued from an architecture with AveragePooling (red curves). It's worth noting here that even the gradients after an AveragePooling contain some artifacts in the high-frequencies  (illustrated by the tip of the curves going up). These artifacts may come from striding mechanisms and carried throughout layers. 

\begin{figure*}[h]
\center
\includegraphics[width=0.99\textwidth]{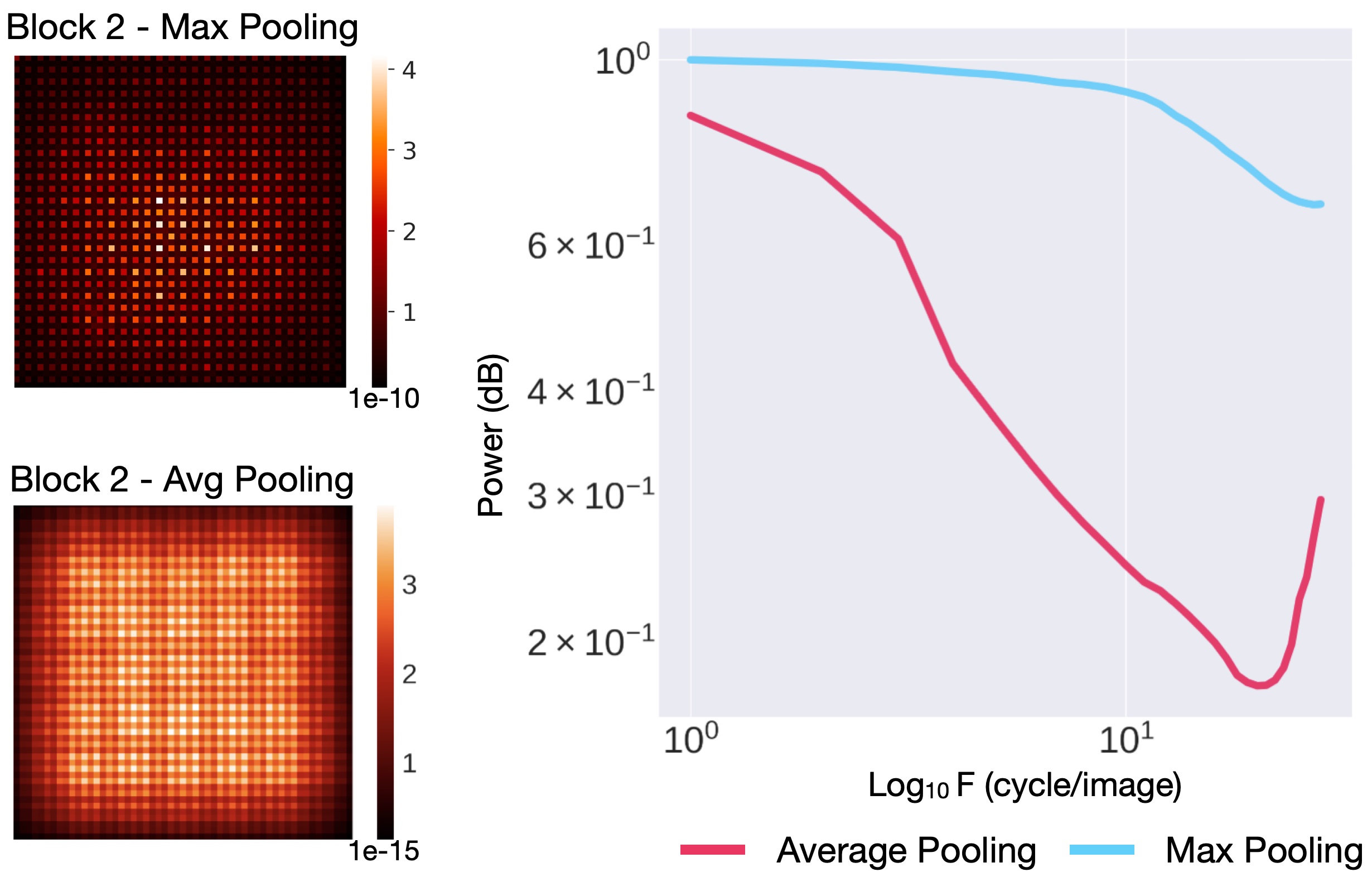}
\caption{{\bf Frequency profiles in every layer of a ResNet50:} Evolution of the power/frequency slope of the gradients in each layer of a ResNet50 network. Here we consider a standard pre-trained ResNet50 (blue curve), an untrained ResNet50 network (black curve), and a ResNet50 in which we substitute MaxPooling with AveragePooling (red curve). The untrained network has the same frequency profile as the trained ResNet50. The ResNet50 with AveragePooling shows a lower power/frequency slope, indicating lower contamination of the high-frequency. All in all, this figure suggests that high frequency in the ResNet50 is not a by-product of the training but rather a consequence of the MaxPooling operation.}
\label{fig_app:resnet_layers}
\end{figure*}

We finally compare the contribution of downsampling operations in the frequency content of the gradients with all the other layers. Fig.~\ref{fig_app:resnet_layers} represents the power/frequency slope of the gradient at each layer, from the prediction to the input. The blue curve represents a pre-trained ResNet50V2 with MaxPooling operation, while the black curve shows the slopes of an untrained network. These two curves show a comparable evolution per layer, confirming our hypothesis of an architectural mechanism, independent from the optimization. However, the red curve, representing the network with AveragePooling as a downsampling operation, shows a significant reduction of the slopes at each layer.
We highlight here the downsampling steps -- separation of each block -- and invite the reader to compare the slope inside each block only. Indeed, the slopes between blocks are not comparable due to a different image size, and therefore a different length of the Fourier signature. The specific contribution of the pooling in Fig.~\ref{fig_app:resnet_layers} corresponds to the first points of each block. The other layers inside each block do not appear to have a significant impact: the slopes remain stable inside each block.    

\begin{figure*}[h]
\center
\includegraphics[width=0.5\textwidth]{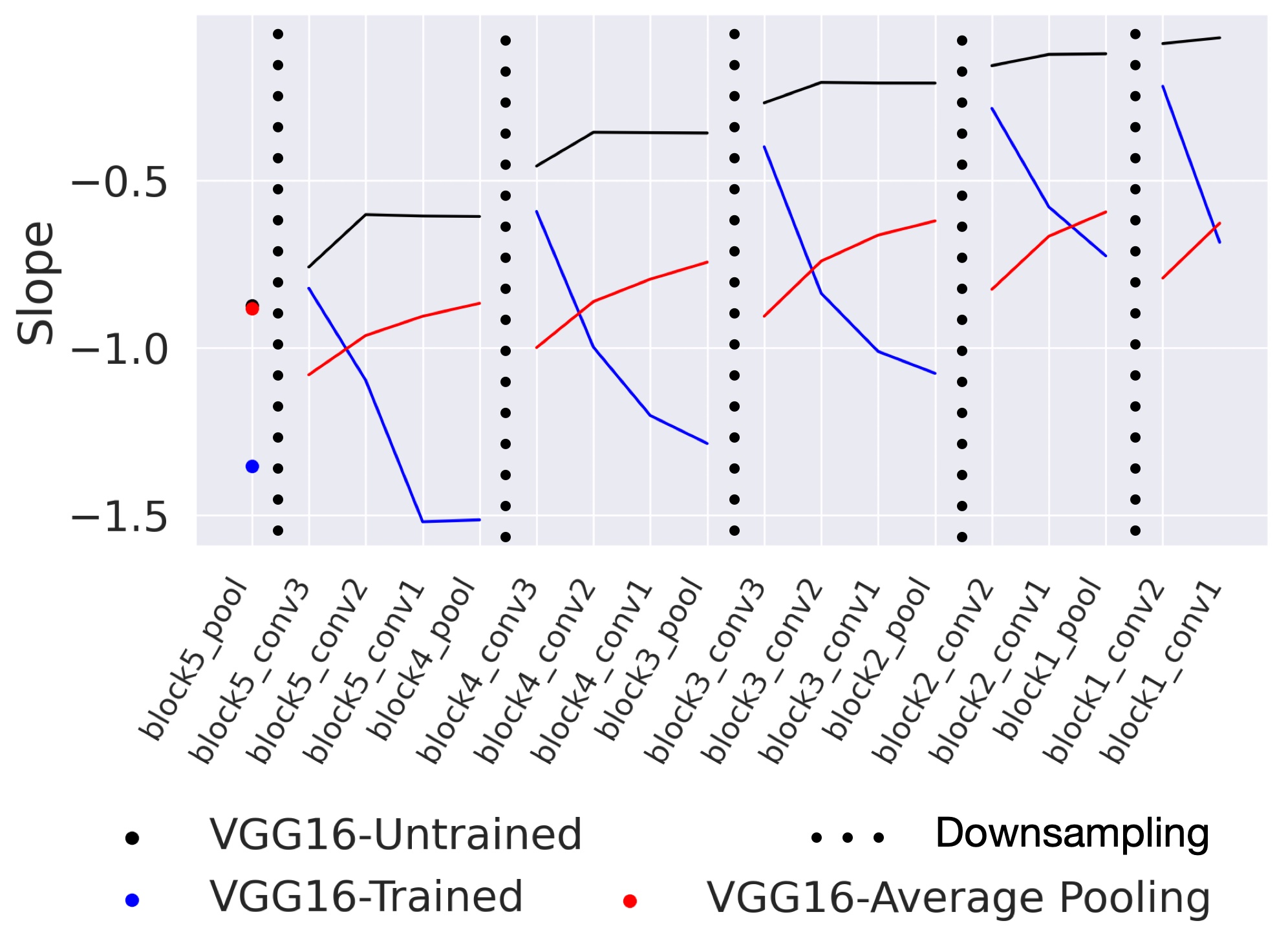}
\caption{{\bf Frequency profiles in every layer of a VGG16:} Evolution of the power/frequency slope of the gradients in each layer of a VGG16 network. Here we consider a standard pre-trained VGG16 (blue curve), an untrained VGG16 network (black curve), and a VGG16 in which we substitute MaxPooling with AveragePooling (red curve). The untrained network has a power/frequency slope slightly higher than the pre-trained version. The VGG16 with AveragePooling shows a lower power/frequency slope at the beginning of each block, indicating lower contamination of the high-frequency. All in all, this figure suggests that high frequency in the VGG16 is not a by-product of the training but rather a consequence of the MaxPooling operation.}
\label{fig_app:vgg_layers}
\end{figure*}

\paragraph{Downsampling operations in a VGG16.} Due to the large number of layers in a ResNet50V2, we perform the same analysis as above on a VGG16 to have a better understanding of the Fourier footprint of the downsampling operations. Fig.~\ref{fig_app:vgg_layers} reproduces the results obtained with ResNet50V2 (Fig.~\ref{fig_app:resnet_layers}. Compared to the above, the operations other than the pooling seem to smooth the gradient (the slopes of the trained VGG16 -- blue curve -- decrease inside each block). However, we still observe a big gap at the beginning of each block due to the MaxPoolings.

\begin{figure*}[h]
\center
\includegraphics[width=0.99\textwidth]{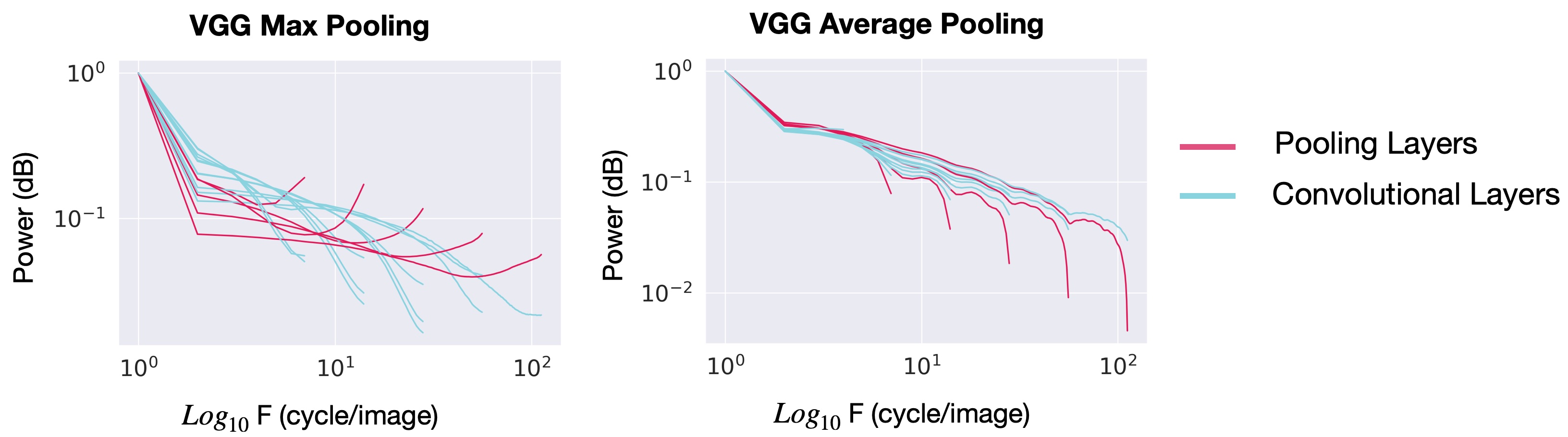}
\caption{{\bf Fourier signature of the gradient after each layer in a VGG16.} The gradients of a VGG16 after MaxPooling operations (right panel -- red curves) show a Fourier signature with more power in the high-frequencies compared to convolution layers in the whole architecture. Comparatively, replacing MaxPooling with AveragePooling lowers the power in the high-frequencies, resulting in a signature with lower high-frequency power than convolution operations.}
\label{fig_app:vgg_max_avg}
\end{figure*}

\begin{figure*}[h]
\center
\includegraphics[width=0.99\textwidth]{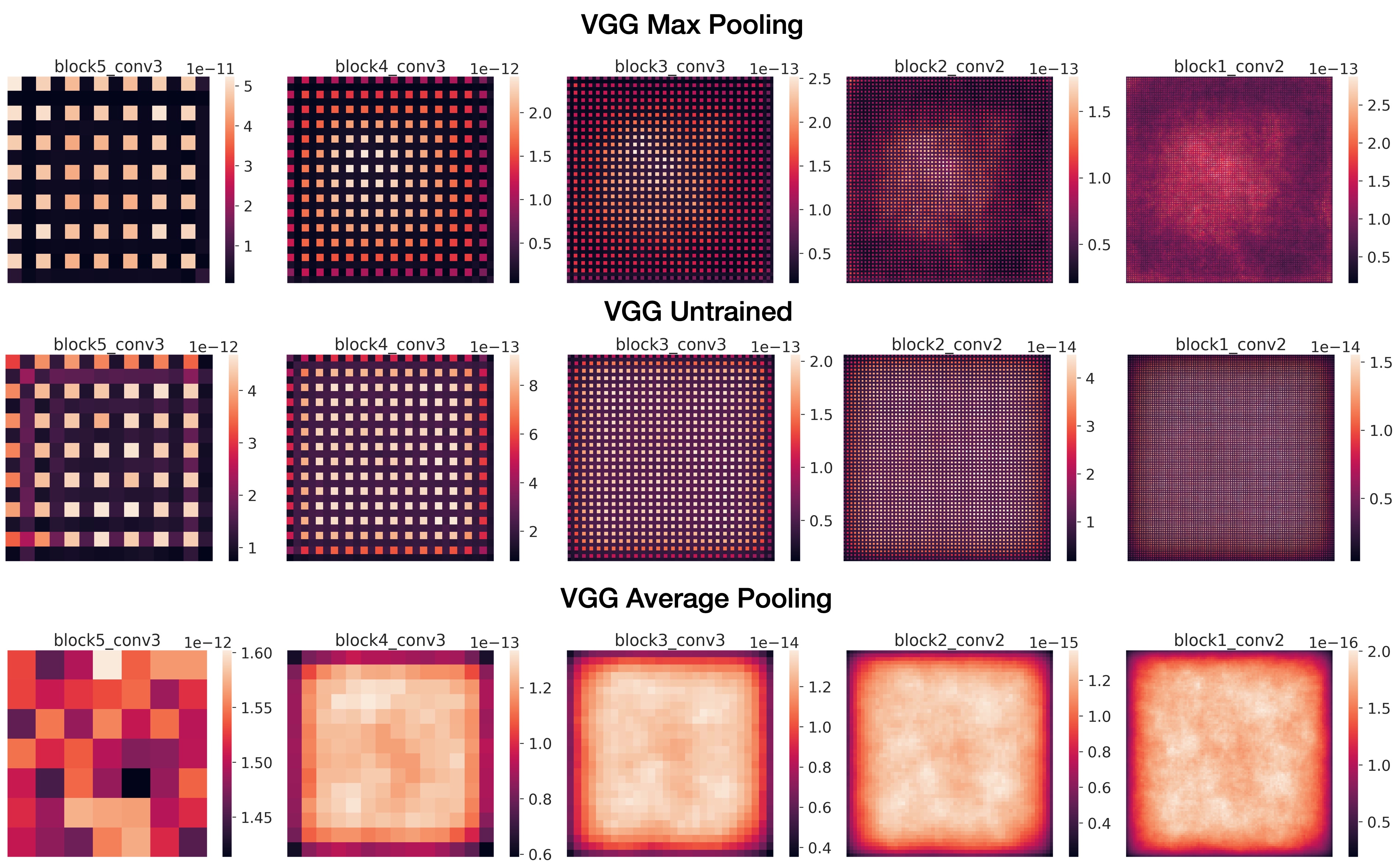}
\caption{Visualization of the gradient (w.r.t. the input) probed at different layers. We observe that both in the VGG MaxPooling and the VGG untrained, a checkerboard signal emerges in the gradients. Such a signal might cause an increased power in high frequencies. In contrast, there is no such signal in the VGG Average Pooling. It suggests that MaxPooling operations, creating such a checkerboard pattern, might contaminate the gradients with high-frequency artifacts.}
\label{fig_app:gradients_vgg}
\end{figure*}

We next plot in Fig.~\ref{fig_app:vgg_max_avg} the Fourier signature of each layer of a VGG16 with MaxPooling (left panel) or AveragePooling (right panel). On both panels, the red curves represent the Fourier signature of the gradient after a pooling operation and the light blue curves correspond to convolutions. On the right panel, we observe a clear difference between the Fourier signature of the MaxPooling compared to the convolutions: the MaxPooling generates a Fourier signature with an increasing power in the higher frequencies. Conversely, the AveragePooling actively decreases the power in these specific high-frequencies, generating Fourier signatures with a significantly reduced frequency power compared to convolutions. The visualization of these gradients at each block, Fig.~\ref{fig_app:gradients_vgg} confirms our quantitative analysis.

\begin{figure*}[h]
\center
\includegraphics[width=0.99\textwidth]{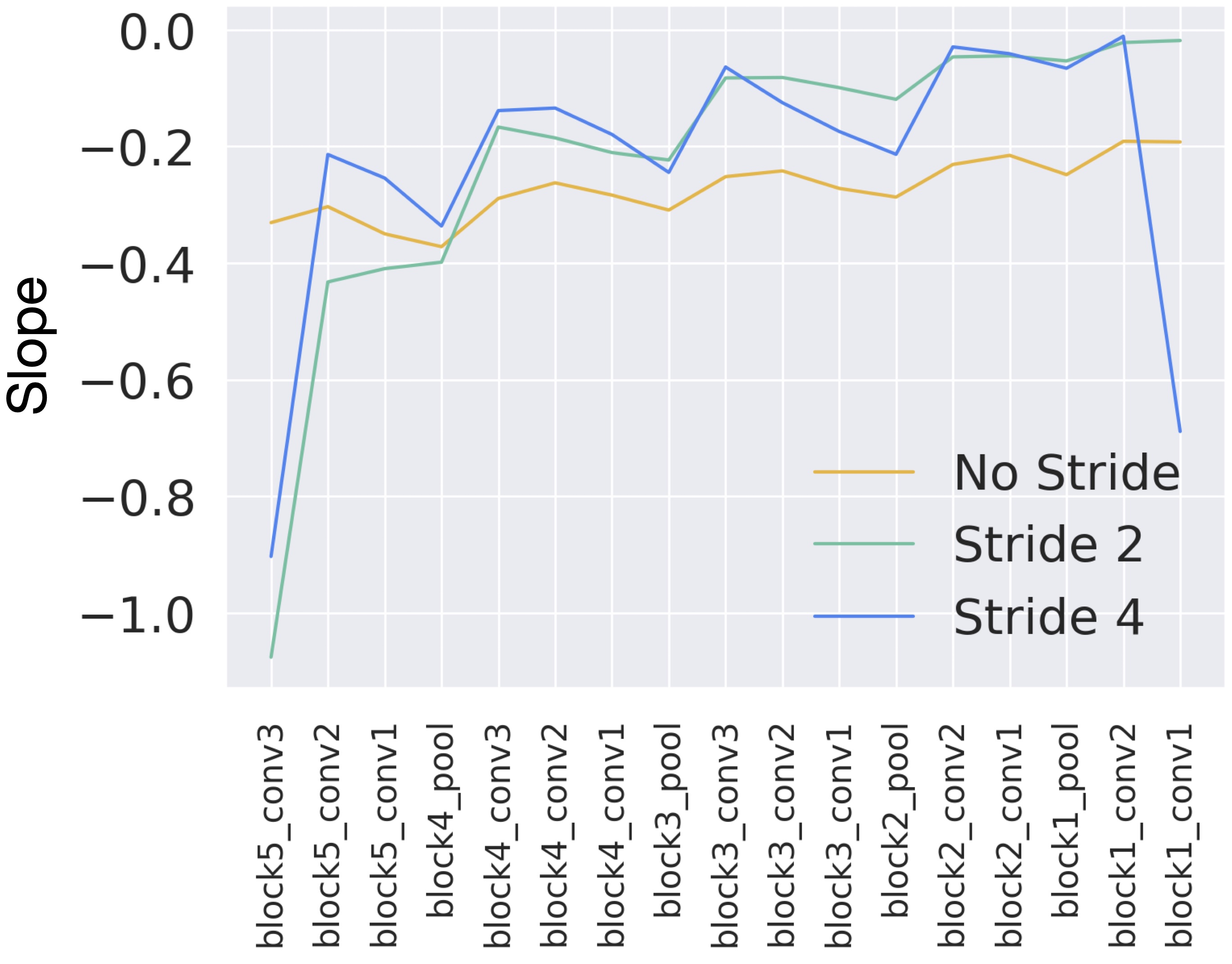}
\caption{Effect of the stride value of MaxPooling in the VGG16. We observe that performing a MaxPooling with a stride of 1 (akin to no stride), reduces the amount of high-frequency artifacts included in the gradients of the model. Varying the stride value (2 or 4) does not change the high-frequency power.}
\label{fig_app:strides_vgg}
\end{figure*}

Finally, we vary the stride value of the MaxPooling operations in a trained VGG16 with MaxPooling2D. In Fig~\ref{fig_app:strides_vgg}, we find that performing a MaxPooling with no stride (meaning using a value of 1) reduces the amount of high-frequency content introduced per layer, compared to using a stride value superior to 1.
However, increasing the stride value from 2 to 4 does not change significantly the amount of high-frequency content.

\newpage
\newpage

\subsection{The importance of an adaptive sigma}
\begin{figure*}[h]
\center
\includegraphics[width=0.99\textwidth]{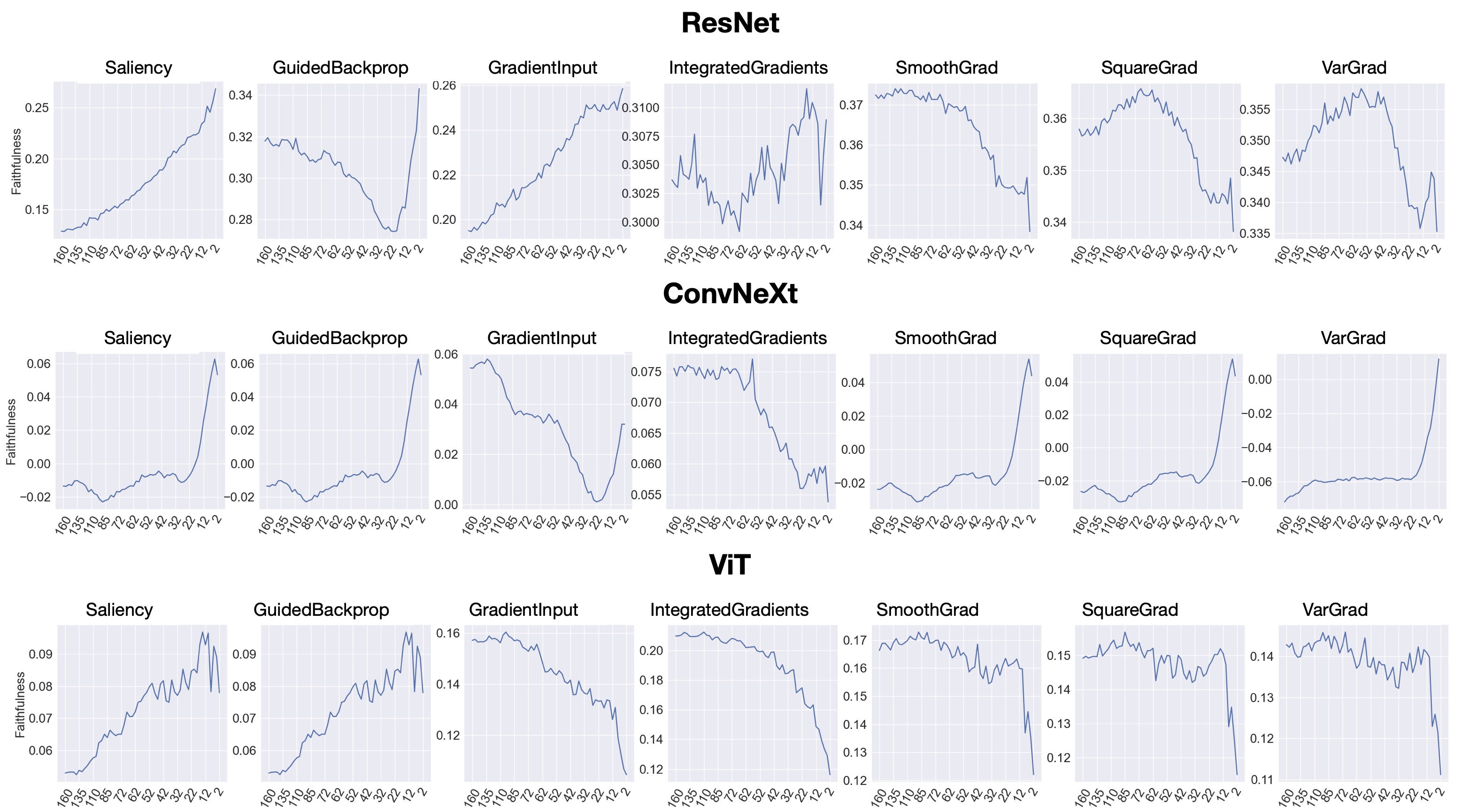}
\caption{{\bf Evolution of the faithfulness depending on the cut-off frequency $\sigma$.} The faithfulness evolves differently per methods and models depending on the cut-off frequency $\sigma$. The filtering tends to be more beneficial for convolutional networks (ResNet50 and ConvNeXt) compared to ViT due to the absence of pooling on the latter family of models. Consistently, filtering improves the faithfulness score or at least does not reduce it.}
\label{fig_app:sigmas_models_methods}
\end{figure*}

\begin{figure*}[h]
\center
\includegraphics[width=0.99\textwidth]{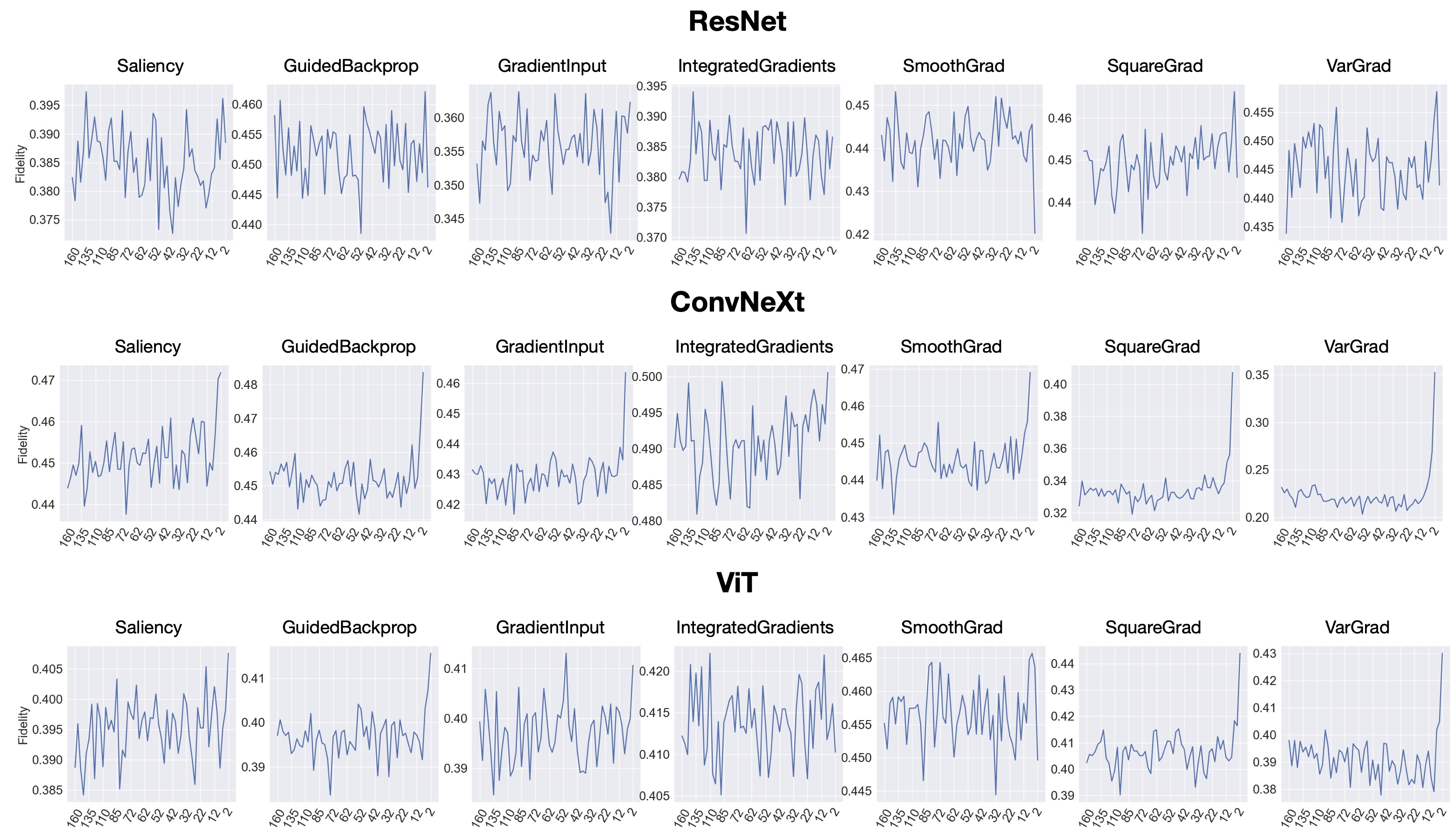}
\caption{{\bf Evolution of the fidelity depending on the cut-off frequency $\sigma$.} The faithfulness evolves differently per methods and models depending on the cut-off frequency $\sigma$. Consistently, filtering improves the faithfulness score or at least does not reduce it.}
\label{fig_app:fidelity_models_methods}
\end{figure*}

In Fig.~\ref{fig_app:sigmas_models_methods}, we show the evolution of the faithfulness w.r.t the value of $\sigma$. We select the optimal cutoff frequency $\sigma\star$ by maximizing the faithfulness. These values range from $224$ to $2$ (we plot the values here starting at $160$ for the sake of visualization). We can observe that in most cases, the faithfulness score increases significantly after filtering the gradient. 
We also observe a pattern of evolution similar for the methods belonging to the same sub-categories (i.e. SmoothGrad, SquareGrad, VarGrad). 
Additionally, the filtering is more useful for convolutional networks (ResNet50V2, ConvNeXt) than for ViT. This can be explained by the absence of MaxPooling or striding in transformer architectures and smoother gradients.
In Fig.~\ref{fig_app:fidelity_models_methods}, we plot the fidelity metric using the same $\sigma$ value as in Fig.~\ref{fig_app:sigmas_models_methods}. The pattern of evolution is significantly different, and the $\sigma$ value maximizing the faithfulness metric does not always correspond to the optimal $\sigma$ value for the fidelity metric. However, it is possible to optimize $\sigma$ on the fidelity metric as there is at least one value per model and method which is increased compared to the absence of filtering.

\newpage
\subsection{Qualitative examples}
\begin{figure*}[h]
\center
\includegraphics[width=0.7\textwidth]{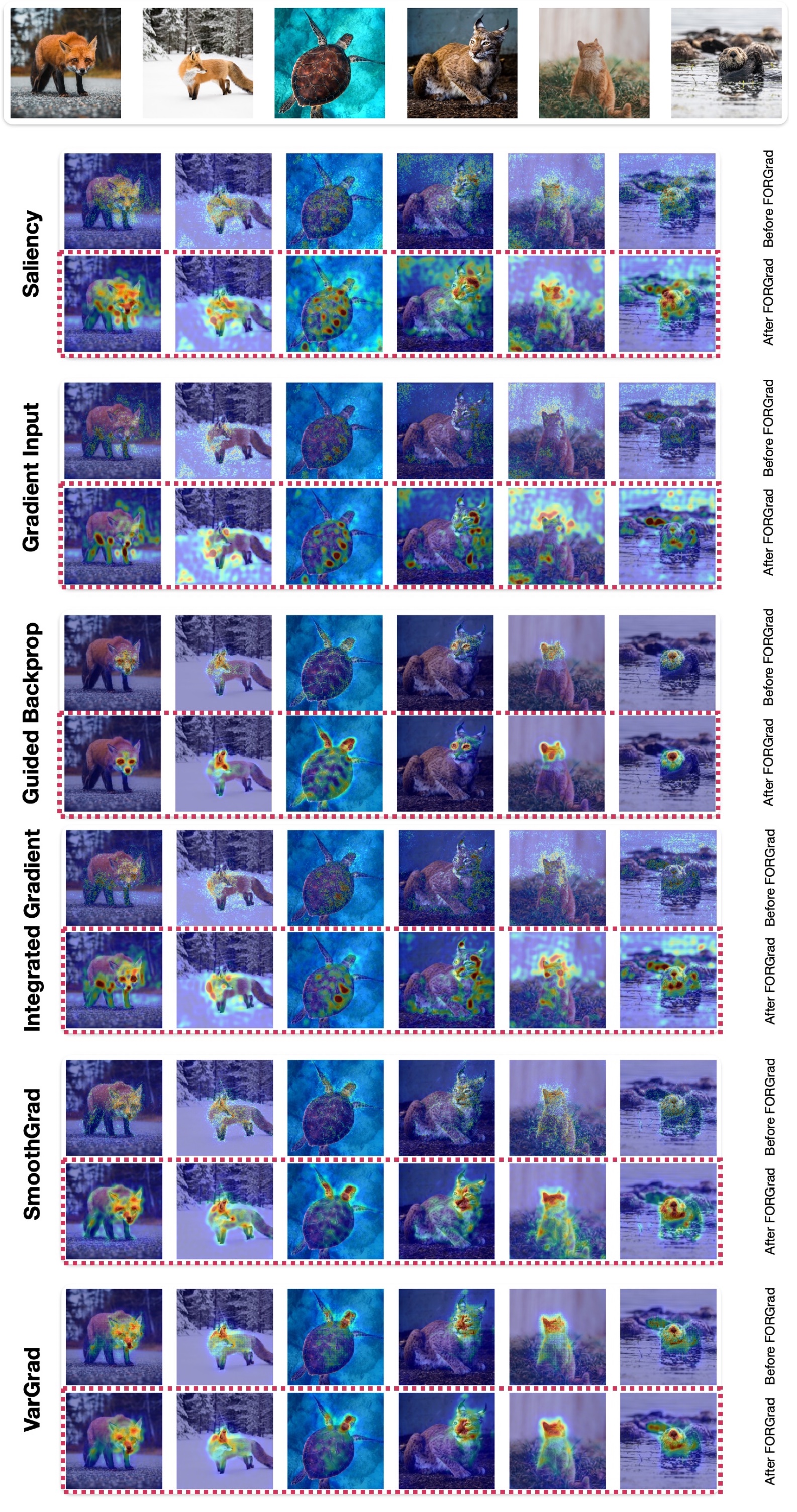}
\caption{{\bf Qualitative examples of several white-box methods before and after \forgrad}}
\label{fig_app:qualitative}
\end{figure*}

We provided in Fig.~\ref{fig_app:qualitative} some qualitative examples of \forgrad. For several White-box methods, we show the produced attribution map before and after applying \forgrad. All the resulting explanations concord with the metrics, showing a clear improvement in quality, and smoother attribution maps. 

\newpage
\subsection{Control experiments} 
\begin{figure*}[h]
\center
\includegraphics[width=0.99\textwidth]{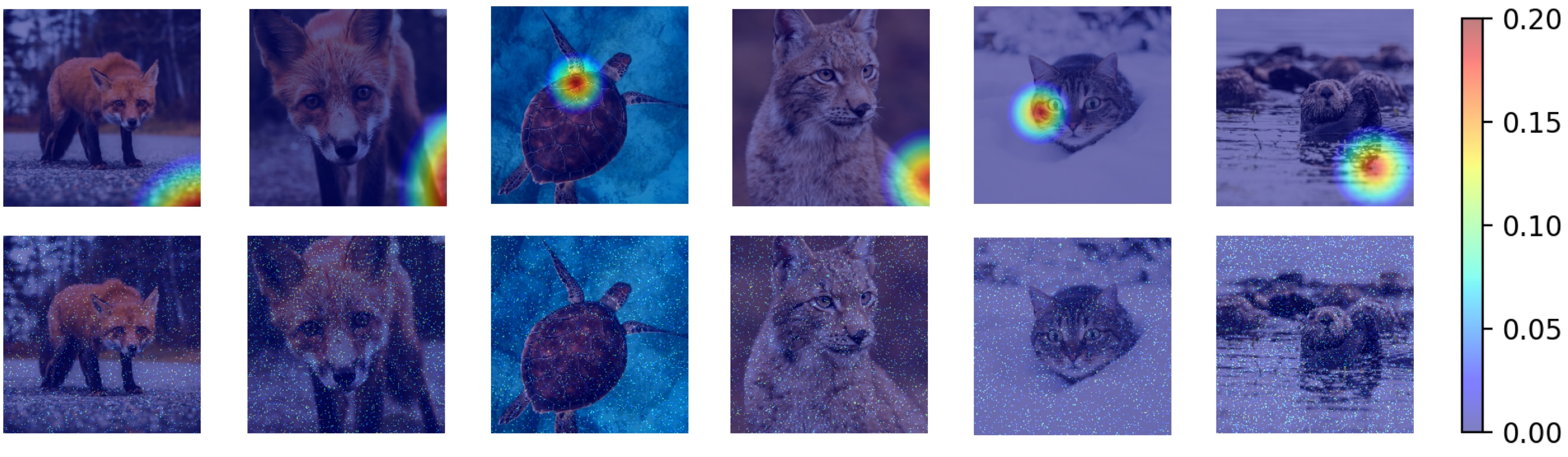}
\caption{{\bf Visualization of low and high-frequency random attribution maps.} The low-frequency random maps consist of 2D Gaussian randomly located in an image (the background images are for the sake of visualization). The high-frequency equivalent consists of dispatching the pixels of the Gaussian anywhere on the image.}
\label{fig_app:random_exp}
\end{figure*}

We provide in this section some control experiments on \forgrad.
\paragraph{Controlling the metrics.} As we use two main metrics to compute our quantitative results, we perform control experiments ensuring none of the metrics are biased towards attribution maps containing low or high-frequencies. To do so, we produce random explanations of very low frequencies by drawing a 2D Gaussian at a random position in an empty map. We generate the high-frequency equivalent by dispatching all the active pixels of the low-frequency map in another empty map. Some examples are shown in Fig.~\ref{fig_app:random_exp}. We then compute the scores given by each metric for 1000 maps of each category. The results are reported in Tab~\ref{tab:control_random}. We add the scores obtained with Saliency as a baseline. The results show a potential bias towards high-frequency explanations explaining why the scores obtained with \forgrad~are a bit less salient. However, no bias appears with faithfulness, confirming the real improvement of \forgrad~on each of the white-box methods. 

\begin{table}
    \centering
    \begin{tabular}{|c|c c c c|} \hline 
         Method name&  Deletion &  Insertion&  Faithfulness& $\mu$ Fidelity\\ \hline 
         Saliency&  0.127&  0.316&  0.189& 0.390\\ 
         Random low-freq.&  0.205&  0.187&  -0.018& 0.140\\ 
         Random high-freq.&  0.138&  0.179&  0.041& 0.310\\ \hline
    \end{tabular}
    \caption{{\bf Faithfulness and $\mu$Fidelity scores on random explanation of high or low frequencies.}We generate random explanation containing either high or low-frequency information and compute their scores given by each of our two metrics. The faithfulness metric shows no apparent bias while $\mu$Fidelity tends to favor high-frequency explanations.}
    \label{tab:control_random}
\end{table}

\begin{figure*}[h]
\center
\includegraphics[width=0.99\textwidth]{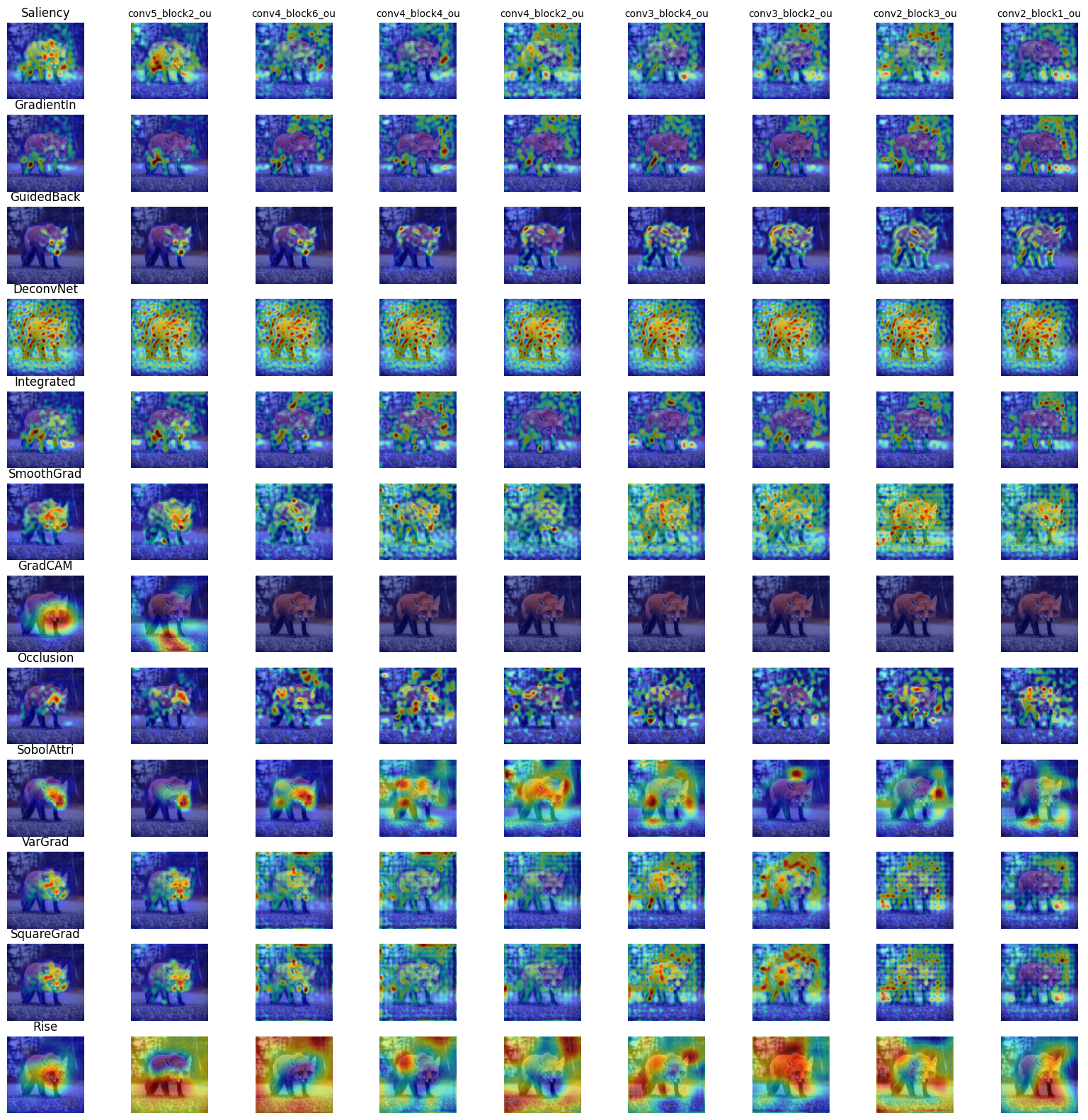}
\caption{{\bf Sanity check of the methods after FORGrad.} We perform the sanity check for attribution methods as presented in \citep{adebayo2018sanity} on a ResNet50. The first column corresponds
to the original heatmap generated by each method. The other columns show the heatmap produced
after randomizing the weights at various layers of the model. Methods that consistently offer a comparable explanation to the original one are deemed to fail the test.}
\label{fig_app:sanity_check}
\end{figure*}

\paragraph{Sanity check for Saliency.} The second experiment reproduces the sanity check proposed by \citet{adebayo2018sanity}. This experiment consists of progressively randomizing the weights of the network and recomputing the attribution map. As a result, the methods that are altered by the randomization pass the test, while the methods producing the same result before or after randomization fail. In Fig.~\ref{fig_app:sanity_check}, we show that \forgrad~ does not change the results of this sanity check. In other words, the methods that failed before still do, and conversely, the ones that passed the test are still affected by the randomization.

\paragraph{Statistical effect of \forgrad.}\label{stats} To show the statistical effect of \forgrad~ for each metric we performed a Bayesian ANOVA (see details below). We analyzed the metrics by means of Bayesian ANOVAs, considering BEFORE-AFTER and ATTRIBUTION-METHODS as fixed factors, and MODELS as a random term. In all analyses, we computed Bayes Factors (BF) as the ratio between the models testing the alternative against the null hypothesis. The alternative hypothesis states that the factor is important for explaining the variance in the data (i.e., there is a significant difference between conditions, for example before and after). All BFs are denoted as BF10. In practice, BFs provide substantial (BF>~5) or strong (BF>~10) evidence in favor of the alternative hypothesis, and low BF (BF<~0.5) suggests a lack of effect \citep{masson2011tutorial}.
Confirming our results, we found a strong difference in the BEFORE-AFTER factor (comparing scores before and after FORGrad) in the INS metric (BEFORE-AFTER factor BF10=11.33, error= 1\%), and moderate evidence for the DEL metric (BEFORE-AFTER factor BF10=1.79, error= 2\%). All in all, considering the small number of samples used for this analysis, we are very confident that the results of the statistical analysis confirm a significant difference between the scores obtained before and after applying FORGrad on the different attribution methods.

\paragraph{Standard Deviation.}
We additionally include in Table~\ref{tab:metrics_std} the standard deviation computed on the scores obtained on each image, for each attribution and metric on ResNet50.

\newpage
\subsection{Quantitative results of \forgrad}
\paragraph{Illustrating the contribution of \forgrad.}We summarize here our contribution with \forgrad. We notice a difference in the Fourier spectra of the different categories of attribution methods. We demonstrate that this difference -- white-box methods show more power in the high-frequencies -- is the reason why black and gray-box methods outperform the first category of methods. The observation is illustrated in the following figure (Fig.~\ref{fig_app:appendix_summary}): white-box methods are mainly located at the right bottom of the plots. \forgrad~ ``repairs'' these methods by filtering the high-frequency artifacts in the gradient, bringing them to the upper-left corner of the plots. 

\begin{figure*}[h]
\center
\includegraphics[width=0.99\textwidth]{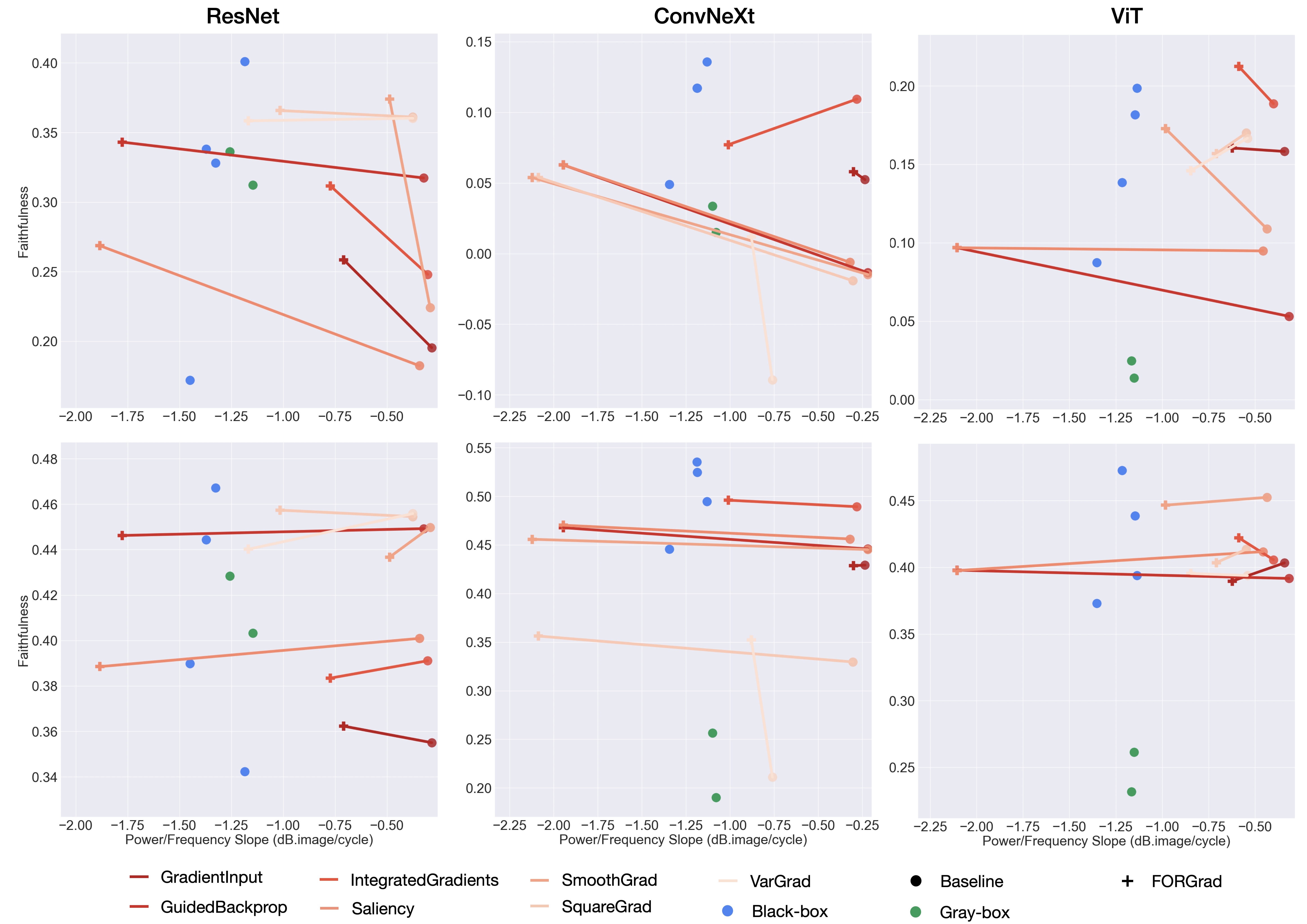}
\caption{{\bf Effect of \forgrad~on white-box attribution methods.} Before applying \forgrad~ most white-box methods (reddish points) contain a lot of power in the high-frequencies (x-axis) but a lower explainability score (y-axis) than other types of methods (blue and green points). \forgrad~corrects white-box methods by low-pass filtering the gradient (reducing the high-frequency power), bringing them back to a faithfulness closer to gray and black-box methods.}
\label{fig_app:appendix_summary}
\end{figure*}

\paragraph{Quantitative evaluation of attribution methods on ResNet50, VIT and ConvNext.}
We additionally report here the faithfulness and fidelity scores of white-box methods before and after \forgrad~ and compare them with gray and black-box methods for ResNet50V2, ConvNeXt, and ViT.
GradCAM methods can't be tested on ViT because they are based on convolution so are limited to CNNs.
For the $3$ models, before \forgrad~ the best methods are always black-box. Even though \forgrad~does not allow to systematically beat them, it brings white-box methods back to competitive performance. Considering their computational efficiency, \forgrad~rends white-box methods very attractive given the accuracy/efficiency tradeoff.

\begin{table*}[t]
    \centering
    \scalebox{0.7}{ %
        \begin{tabular}{l l p{0mm} p{0mm} p{8mm}p{8mm}p{8mm}p{8mm} p{1mm} p{8mm}p{8mm}p{10mm}p{8mm} p{1mm} p{8mm}p{8mm}p{8mm}p{10mm}}
        \toprule
        &&&& \multicolumn{4}{c}{ResNet50} &&  \multicolumn{4}{c}{ConvNeXT} &&  \multicolumn{4}{c}{ViT}  \\
        \cmidrule(lr){5-8} \cmidrule(lr){10-13} \cmidrule(lr){15-18}
        
        &&&& Faith.($\uparrow$) & $\mu$Fid.($\uparrow$) & Stab.($\downarrow$) & Time($\downarrow$)
        && Faith.($\uparrow$) & $\mu$Fid.($\uparrow$) & Stab.($\downarrow$) & Time($\downarrow$)
        && Faith.($\uparrow$) & $\mu$Fid.($\uparrow$) & Stab.($\downarrow$) & Time($\downarrow$)
        \\
        
        \midrule
        \parbox[t]{2mm}{\multirow{16}{*}{\rotatebox[origin=c]{90}{Gradient-based}}}& Saliency\cite{simonyan2013deep} &   
                 && 0.18 & 0.40 & 0.67 & 0.78
                 && -0.01 & 0.46 & 7e-05 & \underline{1.89}
                 && 0.09 & 0.41 & 0.06 & {\bf 0.45} \\
        & Saliency$^\star$ &   
                 && 0.27 & 0.39 & 0.53 & 0.89 %
                 && 0.06 & 0.47 & 3e-05 & {\bf 1.64} %
                 && 0.10 & 0.40 & 0.03 & \underline{0.46} \\ %
        \cmidrule(lr){2-18}
        & Guidedbackprop\cite{ancona2017better} &
                 && 0.31 & 0.45 & 0.28 & 8.25
                 && -0.02 & 0.45 & 4e-05 & 19.3
                 && 0.05 & 0.39 & 0.04 & 5.48 \\
        & Guidedbackprop$\star$ &
                 && 0.34 & 0.45 & 0.22 & 7.05 %
                 && 0.06 & 0.47 & 3e-05 & 17.8  %
                 && 0.10 & 0.40 & 0.03 & 6.41 \\ %
        \cmidrule(lr){2-18}
        & GradInput\cite{ancona2017better} &
                 && 0.20 & 0.36 & 0.42 & {\bf 0.73} 
                 && 0.05 & 0.43 & 0.005 & 1.93
                 && 0.16 & 0.40 & 0.02 & 0.53 \\
        & GradInput$\star$ &
                 && 0.26 & 0.36 & 0.35 & \underline{0.77} %
                 && 0.06 & 0.43 & 0.001 & 1.93  %
                 && 0.16 & 0.39 & 0.006 & 0.59 \\ %
        \cmidrule(lr){2-18}
        & Int.Grad\cite{sundararajan2017axiomatic} &   
                 && 0.24 & 0.39 & 0.72 & 42.7
                 && 0.11 & 0.49 & 0.16 & 93.5
                 && 0.19 & 0.41 & 0.12 & 24.6 \\
        & Int.Grad$\star$ &   
                 && 0.31 & 0.38 & 0.76 & 41.8 %
                 && 0.07 & 0.50 & 0.14 & 93.2  %
                 && {\bf 0.21} & 0.42 & 0.12 & 23.4 \\  %
        \cmidrule(lr){2-18}
        & SmoothGrad\cite{smilkov2017smoothgrad} &   
                 && 0.23 & 0.45 & 0.22 & 46.6
                 && 0.02 & 0.45 & 4e-05 & 104
                 && 0.11 & \underline{0.45} & 0.03 & 26.4 \\
        & SmoothGrad$\star$ &   
                 && \underline{0.37} & 0.44 & 0.21 & 48.3 %
                 && 0.05 & 0.46 & 3e-05 & 103 %
                 && 0.17 & \underline{0.45} & 0.05 & 26.9 \\ %
        \cmidrule(lr){2-18}
        & VarGrad\cite{seo2018noise} &   
                 && 0.36 & \underline{0.46} & {\bf 0.003} & 41.5
                 && -0.09 & 0.21 & 1e-10 & 93.7
                 && 0.16 & 0.39 & 0.001 & 27.1 \\
        & VarGrad$\star$ &   
                 && 0.35 & 0.44 & \underline{0.004} & 40.6 %
                 && 0.01 & 0.35 & 4e-13 & 90.6 %
                 && 0.15 & 0.40 & 0.0003 & 23.4 \\ %
        \cmidrule(lr){2-18}
        & SquareGrad\cite{seo2018noise} &   
                 && 0.36 & 0.45 & {\bf 0.003} & 42.1
                 && -0.02 & 0.33 & 2e-09& 93.4
                 && 0.17 & 0.41 & 0.001 & 20.5 \\
        & SquareGrad$\star$ &   
                 && 0.36 & \underline{0.46} & 0.005 & 40.9 %
                 && 0.05 & 0.36 & 4e-10 & 90.7 %
                 && 0.17 & 0.41 & 0.001 & 19.3 \\ %
        \midrule
        \midrule
        \parbox[t]{2mm}{\multirow{6}{*}{\rotatebox[origin=c]{90}{Prediction-based}}} & GradCAM\cite{Selvaraju_2019} &   
                 && 0.31 & 0.40 & 0.31 & 5.24
                 && 0.03 & 0.26 & 0.005 & 6.80
                 && n.a & n.a & n.a & n.a \\
        & GradCAM++\cite{chattopadhay2018grad} &
                 && 0.33 & 0.43 & 0.34 & 4.61
                 && 0.02 & 0.19 & 0.001 & 7.06
                 && n.a & n.a & n.a & n.a \\
        & Occlusion\cite{ancona2017better} &   
                 && 0.17 & 0.39 & 0.60 & 368
                 && 0.05 & 0.45 & 0.05 & 782
                 && 0.09 & 0.37 & 0.09 & 183 \\
        & HSIC\cite{novello2022making}     &   
                 && 0.33 & {\bf 0.47} & 0.45 & 456
                 && 0.08 & {\bf 0.54} & 0.14 & 1358
                 && \underline{0.20} & {\bf 0.47} & 0.14 & 248 \\
        & Sobol\cite{fel2021sobol}     &   
                 && 0.34 & {\bf 0.47} & 0.47 & 578
                 && \underline{0.12} & \underline{0.53} & 0.16 & 1156
                 && 0.14 & {\bf 0.47} & 0.05 & 342 \\
        & RISE\cite{petsiuk2018rise}     &   
                 && {\bf 0.41} & 0.34 & 0.55 & 626
                 && {\bf 0.13} & 0.49 & 0.15 & 1534
                 && 0.18 & 0.44 & 0.08 & 487 \\
        \bottomrule \\
    \end{tabular}
    }
    \caption{{\bf Results on Faithfulness metrics}. Faithfulness and Fidelity scores were obtained on 1,280 ImageNet validation set images, on an Nvidia T4 (For Faithfulness and Fidelity, higher is better). Time in seconds corresponds to the generation of 100 (ImageNet) explanations on an Nvidia T4.
    The first and second best results are in \textbf{bold} and \underline{underlined}.}%
    \label{tab:fathifulness_full}
    \vspace{-5mm}
\end{table*}

\begin{table*}[t]
    \centering
    \scalebox{0.70}{ %
        \begin{tabular}{l cc c cc c cc}
        \toprule
        & \multicolumn{2}{c}{ResNet50} &&  \multicolumn{2}{c}{ConvNeXT} &&  \multicolumn{2}{c}{ViT}  \\
        
        & Original & \forgrad && Original & \forgrad && Original & \forgrad
        \\
        
        \midrule
        
         1 & VarGrad (0.82) & \best{SquareGrad$\star$ (0.82)} && Sobol (0.49) & \best{Saliency$\star$ (0.53)} && SquareGrad (0.579) & \best{SquareGrad$\star$ (0.579)} \\

         2 & SquareGrad (0.81) & \best{VarGrad$\star$ (0.82)} &&  HSIC (0.48) & \best{GuidedBackprop$\star$ (0.53)} && Sobol (0.56)  & \best{SmoothGrad$\star$ (0.57)} \\

         3 & GuidedBackprop (0.48) & \best{SmoothGrad (0.60)} && GradInput (0.475) & \best{SmoothGrad$\star$ (0.51)} && VarGrad (0.549) & Sobol (0.56) \\
         
         4 & SmoothGrad (0.46) & \best{GuidedBackprop$\star$ (0.58)} && Rise (0.47) & Sobol (0.49) && GradInput (0.54) & \best{VarGrad$\star$ (0.55)} \\

         5 & GradCAM++ (0.42) & GradCAM (0.42) && SmoothGrad (0.47) & \best{GradInput$\star$ (0.489)} && Rise (0.54) & GradInput (0.544) \\

        \bottomrule \\
    \end{tabular}
    }
    \caption{{\bf Global ranking before (original) and after \forgrad.} For each model, we show the 5 attribution methods with the highest metrics, before and after applying \forgrad. The explanation maps were computed on $1280$ images from the validation set of ImageNet, based on an aggregation of the three metrics computed by $F(\x, \am) + \mu F(\x, \am) - S(\x,\am)$.
    }
    \label{tab:ranking_full}
\end{table*}

\paragraph{Additional quantitative control measurements.}
We finally include additional experiments to control the robustness of~\forgrad~on the different metrics and parameters.
Table~\ref{tab:baseline} summarizes the results of~\forgrad~when the baseline value of Insertion and Deletion metrics is changed from 0 to uniform noise between -1 and 1. We can see that the scores indeed differ, but the effectiveness of the method does not.
Table~\ref{tab:metrics_std} evaluates the effect of applying~\forgrad~(i.e. low-pass filtering using an optimal cut-off value) after computing the attribution map instead of filtering the gradient prior to generating the attribution map ($\dagger$ symbol) on ResNet50. Even though the metrics improve compared to the absence of filtering, filtering the gradient before computing the attribution map is the most optimal.

\begin{table}[t]
    \centering
    \scalebox{0.9}{ %
        \begin{tabular}{p{1mm} p{57mm} p{10mm}p{10mm}p{10mm}}
        \toprule
        && \multicolumn{3}{c}{ResNet50} \\
        \cmidrule(lr){3-5} %
        && Del.($\downarrow$) & Ins.($\uparrow$) & Faith.($\uparrow$)
        \\
        \midrule
        \parbox[t]{2mm}{\multirow{16}{*}{\rotatebox[origin=c]{90}{White-box}}}& Saliency\cite{simonyan2013deep} &  
                  0.08 & 0.27 & 0.17 \\ 
        & Saliency$^\star$ &   
                  0.09 & {\bf 0.37} & {\bf 0.28} \\ 
        \cmidrule(lr){2-5}
        & Guidedbackprop\cite{ancona2017better} &
                  0.05 & 0.33 & 0.28 \\
        & Guidedbackprop$\star$ &
                  0.07 & {\bf 0.39} & {\bf 0.32} \\
        \cmidrule(lr){2-5}
        & GradInput\cite{shrikumar2017learning} &
                  0.08 & 0.23 & 0.15 \\ 
        & GradInput$\star$ &
                  0.11 & {\bf 0.33} & {\bf 0.35} \\ 
        \cmidrule(lr){2-5}
        & Int.Grad\cite{sundararajan2017axiomatic} &   
                  0.11 & 0.17 & 0.06 \\
        & Int.Grad$\star$ &   
                  {\bf 0.10} & 0.17 & {\bf 0.07}\\
        \cmidrule(lr){2-5}
        & SmoothGrad\cite{smilkov2017smoothgrad} &   
                  0.13 & 0.31 & 0.18 \\
        & SmoothGrad$\star$ &   
                  0.13 & {\bf 0.34} & {\bf 0.21} \\ %
        \cmidrule(lr){2-5}
        & VarGrad \cite{adebayo2018sanity} &   
                  0.12 & 0.28 & 0.16\\
        & VarGrad$\star$ &   
                  0.13 & {\bf 0.33} & {\bf 0.20} \\ %
        \cmidrule(lr){2-5}
        & SquareGrad\cite{seo2018noise} &   
                  0.12 & 0.29 & 0.17 \\
        & SquareGrad$\star$ &   
                  0.13 & {\bf 0.33} & {\bf 0.20}\\ 
        \midrule
        \midrule
        \parbox[t]{2mm}{\multirow{6}{*}{\rotatebox[origin=c]{90}{Black \& Gray-box}}} & GradCAM\cite{Selvaraju_2019} &   
                  0.10 & 0.40 & 0.30\\
        & GradCAM++\cite{chattopadhay2018grad} &
                  0.08 & 0.41 & 0.33\\
        & Occlusion\cite{ancona2017better} &   
                  0.08 & 0.28 & 0.20\\
        & HSIC\cite{novello2022making}     &   
                  0.06 &  0.42 &  0.36\\
        & Sobol\cite{fel2021sobol}     &   
                  0.07 & 0.40 &  0.33\\
        & RISE\cite{petsiuk2018rise}     &   
                  0.07 & 0.41 &  0.34\\
        \bottomrule \\
            \end{tabular}
    }
    \caption{{\bf Results on Deletion, Insertion, and Faithfulness metrics}. Deletion, Insertion, and Faithfulness scores were obtained on 1,280 ImageNet validation set images, on an Nvidia T4 (For Insertion and Faithfulness, higher is better).
    The scores that are improved by \forgrad~ are in \textbf{bold}}%
    \label{tab:baseline}
    \vspace{-5mm}
\end{table}

\begin{table}[t]
    \centering
    \scalebox{0.9}{ %
        \begin{tabular}{p{1mm} p{57mm} p{20mm}p{20mm}p{23mm}p{8mm}}
        \toprule
        && \multicolumn{4}{c}{ResNet50} \\
        \cmidrule(lr){3-6} %
        && Faith.($\uparrow$) & $\mu$Fid.($\uparrow$) & Stab.($\downarrow$) & Time($\downarrow$)
        \\
        \midrule
        \parbox[t]{2mm}{\multirow{16}{*}{\rotatebox[origin=c]{90}{White-box}}}& Saliency\cite{simonyan2013deep} &   
                  0.18 $(\pm 0.03)$  & 0.40 $(\pm 0.06)$ & 0.67 $(\pm 0.3)$ & 0.78 \\
        & Saliency$^\dagger$ &   
                  0.28 $(\pm 0.03)$ & 0.41 $(\pm 0.06)$ & 0.56 $(\pm 0.23)$ & 0.87 \\ 
        & Saliency$^\star$ &   
                  0.28 $(\pm 0.02)$ & 0.39 $(\pm 0.06)$ & 0.53 $(\pm 0.16)$ & 0.89 \\ 
        \cmidrule(lr){2-6}
        & Guidedbackprop\cite{ancona2017better} &
                  0.31 $(\pm 0.07)$ & 0.45 $(\pm 0.06)$ & 0.28 $(\pm 0.11)$ & 8.25 \\
        & Guidedbackprop$\dagger$ &
                  0.33 $(\pm 0.03)$ & 0.43 $(\pm 0.07)$ & 0.29 $(\pm 0.06)$ & 7.94 \\
        & Guidedbackprop$\star$ &
                  0.35 $(\pm 0.03)$ & 0.45 $(\pm 0.07)$ & 0.22 $(\pm 0.06)$ & 7.05 \\
        \cmidrule(lr){2-6}
        & GradInput\cite{shrikumar2017learning} &
                  0.2 $(\pm 0.04)$ & 0.36 $(\pm 0.06)$ & 0.42 $(\pm 0.2)$ & {\bf 0.73} \\
        & GradInput$\dagger$ &
                  0.22 $(\pm 0.03)$ & 0.37 $(\pm 0.07)$ & 0.39 $(\pm 0.12)$ & 0.75 \\ 
        & GradInput$\star$ &
                  0.26 $(\pm 0.01)$ & 0.36 $(\pm 0.06)$ & 0.35 $(\pm 0.11)$ & \underline{0.77} \\ 
        \cmidrule(lr){2-6}
        & Int.Grad\cite{sundararajan2017axiomatic} &   
                  0.24 $(\pm 0.05)$ & 0.39 $(\pm 0.06)$ & 0.72 $(\pm 0.21)$ & 42.7 \\
        & Int.Grad$\dagger$ &   
                  0.26 $(\pm 0.03)$ & 0.40 $(\pm 0.06)$ & 0.72 $(\pm 0.20)$ & 41.3 \\
        & Int.Grad$\star$ &   
                  0.31 $(\pm 0.03)$ & 0.38 $(\pm 0.06)$ & 0.76 $(\pm 0.20)$ & 41.3 \\
        \cmidrule(lr){2-6}
        & SmoothGrad\cite{smilkov2017smoothgrad} &   
                  0.23 $(\pm 0.04)$ & 0.45 $(\pm 0.07)$ & 0.22 $(\pm 0.06)$ & 46.6 \\
        & SmoothGrad$\dagger$ &   
                  0.32 $(\pm 0.02)$ & 0.43 $(\pm 0.06)$ & 0.22 $(\pm 0.08)$ & 47.0 \\
        & SmoothGrad$\star$ &   
                  \underline{0.37} $(\pm 0.02)$ & 0.44 $(\pm 0.06)$ & 0.21 $(\pm 0.04)$ & 48.3 \\ %
        \cmidrule(lr){2-6}
        & VarGrad \cite{adebayo2018sanity} &   
                  0.36 $(\pm 0.05)$ & \underline{0.46} $(\pm 0.07)$ & {\bf 0.003} $(\pm 0.001)$ & 41.5 \\
        & VarGrad$\dagger$ &   
                  0.34 $(\pm 0.03)$ & 0.44 $(\pm 0.06)$ & 0.009 $(\pm 0.003)$ & 41.1 \\
        & VarGrad$\star$ &   
                  0.35 $(\pm 0.03)$ & 0.44 $(\pm 0.06)$ & \underline{0.004} $(\pm 0.002)$ & 40.6 \\ %
        \cmidrule(lr){2-6}
        & SquareGrad\cite{seo2018noise} &   
                  0.36 $(\pm 0.05)$ & 0.45 $(\pm 0.07)$ & {\bf 0.003} $(\pm 0.001)$& 42.1 \\
        & SquareGrad$\dagger$ &   
                  0.36 $(\pm 0.03)$ & 0.45 $(\pm 0.06)$ & 0.01 $(\pm 0.003)$ & 41.9 \\
        & SquareGrad$\star$ &   
                  0.36 $(\pm 0.03)$ & \underline{0.46} $(\pm 0.06)$ & 0.005 $(\pm 0.003)$  & 40.9 \\ 
        \midrule
        \midrule
        \parbox[t]{2mm}{\multirow{6}{*}{\rotatebox[origin=c]{90}{Black \& Gray-box}}} & GradCAM\cite{Selvaraju_2019} &   
                  0.31 $(\pm 0.04)$ & 0.40 $(\pm 0.06)$ & 0.31 $(\pm 0.13)$ & 5.24 \\
        & GradCAM++\cite{chattopadhay2018grad} &
                  0.33 $(\pm 0.05)$ & 0.43 $(\pm 0.06)$ & 0.34 $(\pm 0.14)$& 4.61 \\
        & Occlusion\cite{ancona2017better} &   
                  0.20 $(\pm 0.03)$ & 0.39 $(\pm 0.06)$ & 0.6 $(\pm 0.25)$ & 368 \\
        & HSIC\cite{novello2022making}     &   
                  0.33 $(\pm 0.04)$ &  {\bf 0.47} $(\pm 0.07)$ &  0.45 $(\pm 0.19)$ & 456 \\
        & Sobol\cite{fel2021sobol}     &   
                  0.34 $(\pm 0.03)$ &  {\bf 0.47} $(\pm 0.05)$ &  0.47 $(\pm 0.21)$ & 578 \\
        & RISE\cite{petsiuk2018rise}     &   
                  {\bf 0.41} $(\pm 0.05)$  &  0.34 $(\pm 0.06)$ &  0.55 $(\pm 0.17)$ & 626 \\
        \bottomrule \\
            \end{tabular}
    }
    \caption{{\bf Results on Faithfulness metrics}. Faithfulness, Fidelity, and Stability scores were obtained on 1,280 ImageNet validation set images, on an Nvidia T4 (For Faithfulness and Fidelity, higher is better).
    Time in seconds corresponds to the generation of 100 (ImageNet) explanations on an Nvidia T4.
    The first and second best results are in \textbf{bold} and \underline{underlined}. $\dagger$ = \forgrad~computed on the attribution map. $\star$ = \forgrad~computed on the gradients.}%
    \label{tab:metrics_std}
    \vspace{-5mm}
\end{table}

\end{document}